\documentclass{ar-1col-S2O}
\usepackage[numbers]{natbib}
\usepackage{url}
\usepackage{amsmath, amsthm, amssymb, bm}
\usepackage{bbold} 
\usepackage{amsfonts}
\usepackage[format=plain,indention=0cm, font=small, labelfont=bf]{caption}
\usepackage{fancyhdr}
\usepackage{enumitem}
\usepackage{tabularx}
\usepackage{siunitx}
\usepackage{ragged2e}
\usepackage[normalem]{ulem}
\usepackage{pgfgantt}
\usepackage{xcolor}
\usepackage{hyperref}
\usepackage[normalem]{ulem}

\definecolor{myyellow}{rgb}{1,0.7,0}
\definecolor{mygreen}{rgb}{0,0.8,0}
\definecolor{barblue}{RGB}{153,204,254}
\definecolor{groupblue}{RGB}{20,80,200}
\definecolor{linkred}{RGB}{165,0,33}

\usepackage[colorinlistoftodos]{todonotes}
\usepackage{comment}

\firstpagenote{Author accepted manuscript. Posted with permission from the
\textit{Annual Review of Control, Robotics, and Autonomous Systems}, Volume 10;
copyright 2027 the author(s), \url{https://www.annualreviews.org}.
Published version: \url{https://doi.org/10.1146/annurev-control-032125-030330}}
\makeatletter
\def\journalinfo{{\jinfofont\@firstpagenote\endgraf}}
\makeatother

\begin{document}

\markboth{Author et al.}{Short title}

\title{Tactile-Based Robotic Manipulation: From Mechanics-Driven to Data-Driven Perspectives}

\author{Jingyi Xiang, Yixuan Wang, and Wenzhen Yuan
\affil{Siebel School of Computing and Data Science, University of Illinois Urbana-Champaign, Urbana, Illinois, USA, 61801; email: yuanwz@illinois.edu}}

\begin{abstract}
Tactile sensing provides robots with direct access to physical interactions at the contact interface, making it a critical modality for robust and dexterous manipulation. Over the past several decades, tactile-based manipulation research has evolved from mechanics-driven approaches, which explicitly interpret tactile signals through contact mechanics and task-specific reasoning, to data-driven approaches that learn tactile representations and manipulation policies directly from data. This paper reviews the development of tactile-based manipulation from these two complementary perspectives. We first introduce the fundamentals of tactile sensing, and then review, from a mechanics-driven perspective, how those signals form manipulation-relevant state representations and subsequently influence the robot action. We then examine the recent advances in data-driven tactile manipulation, covering imitation learning, reinforcement learning, vision-language-action models, and predictive models. Finally, we discuss open challenges and future opportunities toward scalable, generalizable, and robust tactile-enabled robotic systems.
\end{abstract}

\maketitle

\section{INTRODUCTION}

Manipulation is the process by which a robot physically interacts with and changes the state of objects in its environment. Because these interactions occur through contact, many manipulation-relevant quantities—including contact state, force distribution, slip, and local object geometry—can only be observed at the contact interface. Tactile sensing provides direct measurements of these interactions and therefore plays a critical role in enabling robust and dexterous manipulation.

Research on robotic tactile sensing and manipulation has existed for decades. A central question throughout this history has been how tactile signals can be transformed into actionable information for robot control. Early research largely approached this problem from a mechanics-driven perspective, seeking explicit connections between tactile measurements, description of physical interactions, and manipulation actions. For example, tactile feedback can be used to detect slip so as to secure grasp, guide insertion and assembly based on force feedback, or measure local geometry for contact alignment. Although different approaches have been proposed—including analytical models, handcrafted feature extraction, and supervised learning techniques—the common principle is the use of explicit knowledge about contact mechanics and task structure to interpret tactile signals and guide robot behavior. In this review, we collectively refer to these approaches as \emph{mechanics-driven tactile-based manipulation}.

In recent years, the rapid progress of data-driven robot learning has profoundly reshaped tactile-based manipulation research. In contrast to mechanics-driven approaches, these methods typically do not require explicit modeling of intermediate physical quantities. Instead, they learn task-relevant implicit representations and action policies directly from data, often through imitation learning, reinforcement learning, or large-scale multimodal pretraining. This trend emerged alongside the remarkable success of vision-based robot learning, where end-to-end learning frameworks have demonstrated impressive performance across a wide range of manipulation tasks. Consequently, a major research direction in tactile-based manipulation has become the integration of tactile observations into learning architectures originally developed for vision. This direction is particularly promising because vision and touch provide highly complementary information: vision captures global scene context and long-range spatial relationships, while tactile sensing provides precise local measurements of contact and physical interaction. Recent advances in sensor development, scalable robot learning, and multimodal foundation models have further accelerated this trend, creating new opportunities for tactile-enabled manipulation while also introducing new challenges.

In this review, we examine tactile-based manipulation from both mechanics-driven and data-driven perspectives. We first discuss how tactile sensing captures the physics of contact and how tactile signals can be transformed into manipulation-relevant information through explicit reasoning. We then review recent advances in learning-based manipulation, including imitation learning, reinforcement learning, vision-language-action models, and predictive models that leverage tactile observations as part of large-scale multimodal learning frameworks. By bringing together these two perspectives, we aim to provide a unified understanding of how tactile information contributes to robotic manipulation, identify the connections between classical and modern approaches, and highlight the key challenges and opportunities that will shape the future of tactile-enabled robotics. Owing to the breadth of the field, we emphasize representative works that illustrate major ideas and trends rather than providing an exhaustive catalog of existing research.

\section{TACTILE SENSING FUNDAMENTALS} \label{sec:tactile sensor section}

\subsection{Tactile Sensors and Signals}\label{Sec:Sensors}

The first step towards tactile-based manipulation is the tactile sensors. Tactile sensors measure touch, either in the form of a single point force/torque (F/T) sensing or the distributed information over the fingertip or an area of skin. Commonly seen output signals from tactile sensors include: 

\begin{itemize}[noitemsep, topsep=0pt, leftmargin=*]
\item \textbf{Normal Force}, the component of contact force perpendicular to the sensor surface. It is the most commonly measured tactile quantity.
\item \textbf{Shear Force} corresponds to tangential loads at the contact interface and plays a critical role in grasping and in-hand manipulation. The capability to measure shear force is much less common, but some modern tactile sensors, including magnetic \cite{tomo2018new,yan2021soft} and vision-based designs \cite{ward2018tactip,yuan2017gelsight}, can estimate local shear distributions.
\item \textbf{Torque} is available for F/T sensors. For distributed tactile sensors, torque can be inferred from the spatial distribution of forces. 
\item \textbf{Contact Geometry} is typically achieved from the spatial distribution of the deformation or measured normal force across the sensing surface. 
\item \textbf{Vibration} in tactile signals contains important information about contact dynamics and is typically generated during sliding motion. In addition to extracting temporal features of tactile signals, some tactile systems explicitly incorporate accelerometers or vibration sensors to capture high-frequency signals.
\item \textbf{Auxiliary Modalities} such as temperature and proximity. They are less commonly used in robotic manipulation and are therefore outside the scope of this review.
\end{itemize}

Over the past decades, a large number of tactile sensors have been developed \cite{dahiya2009tactile}. They vary in characteristics such as sensitivity, spatial resolution, force range, bandwidth, robustness, and manufacturing complexity. One rough way to categorize the sensors is based on the number of their measurement units. So, there are: 

\begin{itemize}[noitemsep, topsep=0pt, leftmargin=*]
\item \textbf{F/T sensors: } only one measuring unit, but can typically measure 6-axis or 3-axis forces and torque. Installed at the wrist or fingertip  
. The force measurement is usually much more precise.
\item \textbf{Array-based tactile sensors: } typically contain 4-100 sensing units to measure normal force with a spatial resolution of several millimeters. So a rough shape estimation is available but not precise shapes or textures. 
\item \textbf{Vision based tactile sensors: }using an embedded camera to capture raw images, so the spatial resolution can be very high. The downside is the bulkiness in size. Some sensors in this category are designed to measure the surface geometry of every image pixel \cite{yuan2017gelsight,lin2023dtact}, while others track the motion of the printed markers \cite{schneiter1984optical,ward2018tactip}, so their spatial resolution depends on the number of markers. 
\end{itemize}

One major challenge in tactile manipulation is the strong dependency on sensor hardware. The quality and availability of tactile signals vary substantially across sensors, often requiring sensor-specific denoising and preprocessing methods, and training data when using machine learning methods. As a result, algorithms and learning models developed for one sensor frequently transfer poorly to another, and the performance of one manipulation system highly depends on the choice of sensors. We will further discuss this issue in Section~\ref{section:5-heterogeneity}. Throughout this review, we therefore focus on reviewing manipulation methodologies based on how they are built on tactile signals rather than individual sensors.

To abstract out the implementation differences, it is useful to adopt a unified physical view of tactile sensing. Tactile sensors consist of two components: a deformable contact medium, typically made of compliant materials such as elastomers or structures filled with air or liquid, and a transducer that converts the deformation of the medium into measurable electrical signals. Transduction mechanisms may be resistive, capacitive, magnetic, piezoelectric, or optical. Based on this description, tactile sensing is fundamentally indirect: rather than measuring contact itself, sensors measure contact-induced deformation so as to infer information about the interaction. Consequently, the differences among tactile sensors can be featured by the design of the contact medium and the choice of transduction mechanism.

\subsection{Tactile Simulation and Closing Sim-to-Real Gaps}\label{Sec:Simulation}

The growing adoption of data-driven manipulation methods has created a strong demand for tactile sensor simulation. Modern learning-based approaches often require large amounts of training data, while collecting tactile data using physical robots is costly and difficult to scale. Simulation, therefore, provides an attractive alternative for generating large datasets and training manipulation models in a safe and highly parallelized manner. It also provides a good platform for quickly evaluating the design of robotic systems at low cost. 

Accurately reproducing tactile sensor outputs while maintaining high simulation speed remains a long-standing challenge.
While existing technologies cannot perfectly reproduce real sensor responses, improving simulation fidelity also comes at the cost of computational complexity. Since simulation efficiency is also critical for large-scale data generation, practical tactile simulators must balance accuracy and speed. Accordingly, current efforts in tactile simulation generally follow two complementary directions: developing more accurate and efficient sensor models, and designing algorithms to mitigate the sim-to-real gap. 

Unlike conventional robot simulation, tactile simulation must model both the contact mechanics between objects and the sensor, as well as the sensor transduction process that converts deformation into measurable signals. Most robotic simulators employ simplified contact models that cannot fully capture the fine-grained contact phenomena required for tactile sensing, while sensor noise, calibration errors, and manufacturing variations further complicate accurate simulation. 
Existing tactile simulators roughly range from physics-based methods \cite{narang2021sim,agarwal2021simulation,33-akinola2025tacsl,dang2026hydroshear} to data-driven\cite{ruppel2019simulation,sferrazza2020learning,ding2020sim} and hybrid approaches\cite{pezzementi2010characterization,si2022taxim}, each offering different trade-offs between fidelity, computational cost, and adaptability to individual sensors. 
In some applications, particularly those that rely primarily on geometric information, the sim-to-real gap can be sufficiently small to permit zero-shot transfer or transfer with only minor calibration \cite{suresh2022shapemap,si2022grasp}. However, tasks involving force estimation, frictional interactions, slip detection, or dynamic contact typically require substantially more accurate modeling.

To mitigate the sim-to-real gap, one common strategy is to reduce the dimensionality of the tactile representation. Rather than relying on high-resolution raw signals, many manipulation models take input of lower-dimensional contact features such as binary contact states, contact locations, estimated contact geometry, or other task-specific abstractions \cite{30-qi2023general,32-yin2023rotating,35-yuan2024robot,45-yang2025anyrotate,63-yin2025learning}. These representations are often easier to simulate consistently and therefore transfer more reliably to real systems. However, this simplification may discard fine-grained tactile information that could otherwise improve manipulation performance.

When higher-fidelity tactile representations are required, researchers have explored techniques to explicitly bridge the domain gap. One approach is domain randomization, where sensor parameters, contact properties, and measurement noise are randomized during training to improve robustness to simulation errors \cite{gomes2021generation,chen2024general,83-noseworthy2025forge}. Another approach is domain adaptation, which employs transfer learning techniques or generative models to align simulated and real tactile observations \cite{church2022tactile,wang2022tacto,jianu2022reducing,lin2022tactile,zhao2023skill}. Despite these advances, reliable sim-to-real transfer remains an open challenge for tactile-based manipulation.

\section{Tactile-Based Manipulation: A Mechanics-Driven Perspective}

A classical paradigm on tactile-based manipulation is to explicitly model the role that tactile sensing plays in the manipulation process. Under this view, tactile information is incorporated through a pipeline of \textit{perception $\rightarrow$ action}. The \textit{perception} module transforms raw tactile measurements into interpretable descriptions of the contact state that reveal task-relevant information. As a result, it defines what aspects of the physical interaction become observable to the robot and can therefore be exploited for manipulation. The \textit{action} module then utilizes these perceptual estimates to guide planning, decision making, and control. 

A defining characteristic of this perspective is the use of explicit and human-defined representations that connect tactile measurements to robot actions. Although researchers have employed analytical models, handcrafted features, and supervised learning techniques, they share a common philosophy: tactile signals are first interpreted through explicit reasoning based on human prior about contact and interaction before being used for manipulation.

\subsection{Perception} \label{sec:Sec3Perception}
A \textit{perception} module takes the raw tactile signals (see Section~\ref{Sec:Sensors}) as input, and outputs a \emph{perceptual descriptor} that is sensor-agnostic and tied to some particular physical meaning. This general perceptual descriptor, however, can reveal rich information about the state of the manipulation process or the object's properties when interpreted in context. Depending on the sensor choices and system design preferences, the perception module can be a straight-in, straight-out function, an analytical modeling process, or a machine learning model. 
We note that tactile perception more broadly encompasses the general understanding of objects and environments. In this review, however, we restrict our focus to perceptual quantities that are directly relevant to manipulation. We introduce the commonly used perceptual descriptors below:

\subsubsection{Contact force/torque}\label{sec:contact force perception}
Force and torque measurements are among the most fundamental outputs of tactile and haptic sensors (Section~\ref{Sec:Sensors}). Although they are low-dimensional, they often provide rich information about the state of physical interaction. Force and torque values may be obtained directly from force/torque sensors or estimated from raw tactile observations.
Importantly, these measurements can characterize both the direct contact between a robot and an object, as well as indirect interactions occurring elsewhere in the manipulation chain. As a result, force and torque measurements have been widely used to infer task-relevant information and guide robot actions. Examples include:
\begin{itemize}[noitemsep, topsep=0pt, leftmargin=*]
\item \textbf{Maintaining proper contact and grasp stability.} Force feedback can be used to regulate contact force, preventing object slip while avoiding excessive force.\cite{romano2011human}. In some tasks, controlling force also enables sliding motions along an object's surface \cite{1-she2021cable}. 
\item \textbf{Manipulating articulated objects.} Articulated objects, such as doors, drawers, or mechanisms with constrained motion, typically move most easily along specific directions. By monitoring interaction forces, robots can identify directions associated with minimal resistance and follow the underlying kinematic constraints of the object \cite{3-zhao2024tac}.
\item \textbf{Indicating reaching the task stage limit.} When a robot reaches the end of a manipulation limit, say at the end of opening or closing a door, or reaching the end of the hole during pegging, a large impedance force can be detected \cite{su2016learning}. 
\item \textbf{Inferring extrinsic contact states.}
Force measurements can reveal contact events between the objects in the grip and the environment. This can infer task-specific state information, like whether a cable is successfully routed around a pillar~\cite{9-wilson2023cable}. A classic example is peg-in-hole insertion, where interaction forces generated by contact between the peg and the hole can indicate alignment errors \cite{VanWyk2018PegInHole}. Note that in some cases, extrinsic contact states are more precisely estimated from higher-dimensional tactile signals, and we will further discuss it in Section ~\ref{sec:extrinsic contact state}.
\item \textbf{Tool-mediated manipulation.} When a robot indirectly manipulates objects through a tool, force and torque measurements provide important clues about the manipulation state \cite{karayiannidis2014online} and even the state of the objects being in contact\cite{yu2016more}. 
\item \textbf{Object handover.} During human--robot handover, characteristic changes in measured force often indicate that the receiving person has established a stable grasp. Robots can use these cues to determine when it is safe to release the object \cite{Eguiluz2017Handover}.
\end{itemize}

Notably, in many cases, the precision of the force and torque measurement is not essential for perception purposes. For example, for the peg-in-hole, even a very imprecise measurement of the direction of the contact force and torque can well guide the robot to adjust the motion to the correct directions\cite{li2017human}. 

\subsubsection{Shape in contact}
Contact geometry is among the basic outputs of many tactile sensors.  
While a large body of geometry-based tactile perception research focuses on object or material recognition, the role of shape measurement in manipulation is often about verifying whether the correct object or object part has been grasped \cite{12-sunil2023visuotactile,pai2023tactofind}, or estimating the contact location on the objects when the objects' shapes are known \cite{suresh2022shapemap,bauza2023tac2pose}. But more commonly, tactile shape measurements are indirectly used for in-hand localization (details in Section \ref{sec:localization}) and to estimate the pose of manipulated objects (details in Section \ref{sec:pose estimation}).

\subsubsection{Contact location}\label{sec:localization}
Based on tactile geometry measurements, a robot can estimate where it is being touched -- either the location of contact on its body or the position of a grasped object within the gripper. For humanoid robots equipped with whole-body tactile skins, contact localization, combined with force measurements, can help detect unexpected interactions and support impedance control \cite{dean2019whole}. For robotic grippers with fingertip sensors, contact localization can reveal undesirable grasp configurations, such as contacts near the finger edge, allowing the robot to adjust the grasp to enlarge the safety margin \cite{1-she2021cable}.

\subsubsection{Object Pose} \label{sec:pose estimation}
By matching measured contact geometry with a known object shape, a robot can estimate the relative pose between the object and the gripper. Combined with robot kinematics, this estimate can be further transformed to the object's pose in the world frame. Tactile pose estimation may target either the global pose of the object~\cite{bauza2023tac2pose} or the pose of local geometric features, such as edges and corners~\cite{huang2024normalflow}. Accurate pose estimation enables manipulation tasks such as:
\begin{itemize}[noitemsep, topsep=0pt, leftmargin=*]
\item \textbf{Precise manipulation.} Improved pose estimates support accurate object placement \cite{li2014localization,bauza2024Simple}.  In addition, the tactile-based pose estimates can be used as a major reference for repositioning objects in a hand ~\cite{5-hogan2020tactile,49-pitz2023dextrous,21-bronars2024texterity}. 
\item \textbf{Edge following.} 
The local orientation of the edge features will provide important guidance for robots when they need to move along the edge \cite{lepora2017exploratory} or deformable objects such as cables and cloth \cite{1-she2021cable,9-wilson2023cable,12-sunil2023visuotactile}.
\end{itemize}

\subsubsection{Slip} \label{sec:slip detection}
Slip refers to relative motion between an object and the robot at the contact interface. During grasping, both translational and rotational slip are major causes of grasp failure. To prevent failure before it occurs, robots often seek to detect \emph{incipient slip}, a transitional state in which partial detachment occurs at the contact surface before global slip develops. Tactile sensing is particularly well-suited for detecting slip and incipient slip.

From a mechanics perspective, slip can be predicted by comparing normal and tangential forces according to Coulomb's friction law \cite{melchiorri2002slip}. In practice, however, precise force measurements and friction coefficients are often unavailable. Instead, researchers typically infer slip from vibration signals \cite{howe1989sensing,yamada1994tactile,su2015force}, object micro-motions \cite{2-dong2017improved}, or local deformation and stretching of the contact surface \cite{ikeda2004grip,yuan2015measurement,khamis2018papillarray,sui2021incipient}. In addition, learning-based methods have been developed to directly predict slip or incipient slip from tactile observations \cite{reinecke2014experimental,meier2016tactile,james2018slip,veiga2018grip}.

In most manipulation tasks, slip serves as a warning signal of grasp instability, prompting the robots to prevent it (see Section \ref{sec:state transition action}).
However, some tasks intentionally exploit controlled slip, such as sliding along an object's surface to achieve stable contact transitions during manipulation \cite{5-hogan2020tactile,1-she2021cable,21-bronars2024texterity}.

\subsubsection{Extrinsic contact state}\label{sec:extrinsic contact state}

Beyond perceiving contacts directly occurring at the sensor surface, tactile sensing can also be used to infer interactions between a grasped object and the external environment. This capability enables robots to reason about contacts that occur away from the sensing location, which is particularly useful in tool use and indirect manipulation tasks. We previously discussed simple examples based on force and torque measurements in Section~\ref{sec:contact force perception}. More generally, robots may seek to estimate the location, force, or mode of external contacts from tactile observations. Mechanics-driven approaches typically combine tactile measurements with estimates of object pose, geometry, robot kinematics, and contact mechanics to infer the external contact state \cite{Hou2020SharedGrasping,4-kim2022active,Oller2024ContactMode}. While effective, these methods often rely on accurate object models and precise physical reasoning, which can limit their scalability to diverse objects and tasks. To reduce this dependence on explicit modeling, researchers have also explored learning-based approaches that directly infer external contact states from tactile observations \cite{molchanov2016contact}.



\subsubsection{Object property}\label{sec:property perception}
Tactile sensing can be used to infer object properties that are important for manipulation. Such information can help robots select appropriate manipulation strategies and tune low-level controllers so as to precisely manipulate objects with unknown or challenging properties. We will further discuss this in Section~\ref{sec:low-level control}. 

Although tactile object perception is itself a broad research topic, several properties are particularly relevant to manipulation, including center of mass \cite{atkeson1985rigid,guo2023estimating,jin2025com}, hardness \cite{su2012compliance_biotac,yuan2017hardness}, surface roughness, friction, and material-specific properties such as liquid viscosity \cite{Huang-RSS-22,luo2024pancake}. In some cases, these properties are not explicitly estimated but are instead implicitly encoded in learned manipulation policies \cite{wang2020swingbot,huang2024kitchen}.
Estimating such properties often requires \emph{interactive perception}, where the robot actively probes an object and interprets the resulting tactile responses. Consequently, perception and action are often tightly coupled, with exploratory motions serving as an integral part of the perception process.

\subsection{Action}

The \textit{action} module takes in the \emph{perceptual descriptor} from the \textit{perception} module, and outputs actions executable by a robot. 
Depending on the task, the resulting actions may range from short-horizon control commands to longer-horizon motion sequences. While the underlying control and decision-making frameworks are often similar to those used in conventional robotic manipulation, tactile feedback provides additional information about the ongoing physical interaction and enables robots to react to contact events in real time. In this subsection, we review the most common strategies for incorporating tactile perception into robot actions.

\subsubsection{Tactile-driven closed-loop control}

For many contact-rich manipulation tasks, robots must react to tactile feedback in real time. A common strategy is to incorporate tactile perceptual descriptors into a closed-loop control framework, where the current contact state is represented by tactile descriptors and compared with a desired state. The resulting error is then used to generate corrective actions, enabling real-time regulation of contact interactions.

Li et al.\cite{Li-tactile-servoing} introduced a representative formulation of this idea. Using an array-based tactile sensor, the authors represented the contact state using contact location, object orientation, and normal force. Actions in the form of the robot’s joint motion command were generated to reduce the error between the current and target tactile states, where inverse kinematics mapped desired tactile changes into end-effector motions. This tactile servoing framework enabled the robot to maintain desired contact states and track object motion during interaction.

By appropriately defining the tactile state, reference state, and control algorithms, this paradigm can be applied to a wide range of manipulation tasks based different choices of tactile descriptors, including:
\begin{itemize}[noitemsep, topsep=0pt]
\item Obstacle avoidance \cite{Jain2013ClutterWhole} based on contact location (Section \ref{sec:localization}).
\item Peg-in-hole insertion \cite{VanWyk2018PegInHole} based on contact force (Section \ref{sec:contact force perception}).
\item Edge and surface following, including cable manipulation \cite{lepora2017exploratory,lepora2019pixels,12-sunil2023visuotactile,1-she2021cable,9-wilson2023cable} based on contact location (Section \ref{sec:localization}).
\item Manipulation of articulated objects \cite{3-zhao2024tac} based on contact force (Section \ref{sec:contact force perception}).
\item Object reorientation within the gripper \cite{21-bronars2024texterity,5-hogan2020tactile} based on object pose (Section \ref{sec:pose estimation}).
\item Object rotation with multi-DoF contact control \cite{li2015visuo} based on contact force (Section \ref{sec:contact force perception}) and contact location (Section \ref{sec:localization}).
\end{itemize}

The perception-to-action mapping can also be realized implicitly. For example, in reinforcement learning (RL), tactile perceptual signals may be used to define reward functions. The learned RL policy is then optimized to drive the robot motion toward desirable tactile states. For instance, \cite{zheng2022autonomous} trained an RL-based policy for robotic page flipping using rewards derived from the similarity between observed tactile signals and human expert demonstrations, eliminating the need for a hand-designed tactile controller. 

\subsubsection{State transition based on tactile perception}\label{sec:state transition action}
Unlike closed-loop tactile control, where actions are continuously adjusted according to tactile feedback, some manipulation tasks use tactile perception to estimate the current manipulation state and trigger transitions to appropriate actions. In this paradigm, tactile information serves as a decision variable that determines how the robot should proceed in the task.

Grasp stabilization and re-grasping are common examples. While tactile sensing is widely used for detecting slip and grasp failure (Section~\ref{sec:slip detection}), many systems employ the feedback to initiate corrective actions. The most common strategy is to increase the gripping force when slip is detected \cite{romano2011human,ajoudani2016reflex,2-dong2017improved,veiga2018grip,veiga2020grip,5-hogan2020tactile,19-zhang2020towards}. More sophisticated approaches adjust the grasp configuration itself. For example, when rotational slip or excessive torque indicates an unstable grasp, the robot may reposition its contact location toward the object's center of mass \cite{kolamuri2021improving}. Learning-based approaches have also been developed to predict improved grasp configurations directly from tactile observations \cite{calandra2018more,hogan2018tactile}.

A related application arises in long-horizon manipulation tasks that consist of multiple stages. In such settings, tactile perception can be used to detect the completion of a subtask or identify the current interaction state. The manipulation planner then transitions to the next stage, potentially invoking a different motion trajectory, controller, or manipulation strategy \cite{su2016learning,9-wilson2023cable}.

\subsubsection{Perception to update low-Level control strategies} \label{sec:low-level control}
Many manipulation tasks involve objects with unknown or challenging physical properties, where the effectiveness of a controller depends strongly on properties such as mass, center of mass, friction, stiffness, or viscosity. In these cases, tactile perception can be used to estimate object properties and subsequently adapt controller parameters or control strategies. We discussed the tactile-based property perception in Section \ref{sec:property perception}, where the results can be inserted in this framework directly. However, in some more challenging cases, the perceived information may be represented implicitly as a latent feature that captures the aspects of object dynamics most relevant to control. Although these representations may not have a direct physical interpretation, they often encode information similar to physical properties and serve the same purpose of informing controller design.

One example is the dynamic manipulation of objects with unknown weight distributions. In \cite{wang2020swingbot}, a robot first performs exploratory actions to collect tactile observations and infer a latent representation of a rod's physical properties. This representation is then used to select controller parameters for swinging the rod to a desired angle. Similar ideas have been applied to manipulating objects with challenging material properties. For example, \cite{huang2024kitchen,luo2024pancake} use tactile exploration to estimate properties of non-Newtonian fluids such as sauces and batter, allowing the robot to adjust control parameters for dispensing motion and achieve precise and consistent pouring behavior.
\cite{zhang2019leveraging} infers food material with properties regarding softness or skin mechanics from contact vibration and force-torque feedback, then sets the stroke-amplitude and downward-displacement parameters of a slicing primitive accordingly. 

\subsection{Discussion}

In this section, we reviewed the mechanics-driven perspective of tactile-based manipulation, organized through a \textit{perception $\rightarrow$ action} framework. 
The performance of those methods varies and depends heavily on the choice of sensor hardware,  application conditions, and the quality of algorithm design. Their main strengths lie in their interpretability, reliability, and efficiency: because the underlying contact representations and reasoning processes are explicitly modeled, it is often possible to understand when and why a method would succeed or fail. Many of those methods are based on analytical models or lightweight processing modules, so they can quickly process tactile data to meet the requirements of low-latency feedback control.  Some methods exhibit strong generalization across objects and tasks when the underlying assumptions about contact interactions remain valid. However, their performance can degrade when contact interactions become highly complex or difficult to model, especially in real-world settings. These limitations have partially motivated the growing interest in data-driven methods.


The reviewed works also demonstrated fundamental properties of tactile sensing. First, tactile sensing provides local and precise measurements of contact, making it particularly effective for monitoring manipulation state and supporting high-frequency feedback control. Second, tactile perception is inherently active and interactive. Unlike passive sensing modalities such as vision, tactile information is acquired through physical interaction and often evolves as the robot acts on the environment. As a result, perception and action are frequently coupled, with exploratory motions serving not only manipulation objectives but also perception objectives. 

Finally, effective manipulation often requires the integration of multiple sensing modalities. Vision and touch provide highly complementary information.
In mechanics-driven methods, these modalities are typically integrated through explicitly designed representations and reasoning pipelines. As we will discuss in the next section, recent data-driven approaches increasingly seek to learn such multimodal representations directly from data.
\section{Tactile-Based Manipulation: A Data-Driven Perspective}

Data-driven methods learn \textit{policies}—functions that map observations directly to actions—from data. It is in contrast to mechanics-driven approaches, where the perception-to-action process is explicitly modeled. Driven by advances in robot learning, computational resources, and large-scale data collection, this paradigm has achieved remarkable success in vision-based manipulation and has increasingly been adopted in tactile manipulation. Beyond applying existing learning frameworks to tactile inputs, recent research has focused on integrating vision and touch to improve performance in contact-rich tasks. In this section, we review major data-driven paradigms for tactile manipulation, including imitation learning, reinforcement learning, vision-language-action systems, and predictive models, with a focus on how tactile information is utilized within these frameworks.

\subsection{Tactile-Augmented Imitation Learning}
\label{section:4-1}

Imitation learning, a supervised learning approach in which the policy learns to imitate the actions an expert demonstrator performed in similar observed situations, has emerged as a promising direction for learning visuomotor control. Popularized by methods such as Diffusion Policy \cite{chi2025diffusion},
which generate a sequence of future actions conditioned on current observations, imitation learning has demonstrated remarkable results in a wide range of traditionally challenging tasks.
These methods have so far been vision-dominant, and integrating tactile sensing into them has become an active research direction. 
With tactile-inclusive demonstration data collected with teleoperation or hand-held devices, tactile-augmented imitation learning has been shown to handle diverse contact-rich tasks, spanning industrial assembly to everyday manipulation. 

A central question in tactile-augmented imitation learning is how tactile signals should be encoded and fused with existing modalities such as vision and proprioception. It has been widely reported across the literature that naive integration strategies, such as applying vision encoders and their pretrained weights on tactile images, or simply treating tactile signals as additional observation inputs to the policy, often do not improve policy performance \cite{71-ForceMimic, 27-xue2025reactive, 47-romero2024eyesight, 56-zhu2025touch, 68-liu2025factr, 80-he2025foar, 90-chen2026multimodal}. In response, researchers have developed many novel integration strategies, which we organize and discuss below.

\subsubsection{Novel tactile representations} \label{4-1-1}
A representation is a learned or designed encoding that maps raw, high-dimensional sensory inputs into a compact feature space that captures its task-relevant structure. 
Suitable representations can greatly improve the effectiveness of downstream policy learning \cite{VINN}.
Developing novel tactile representations is especially active for high-dimensional tactile signals that often get treated as images, such as those from tactile arrays and vision-based tactile sensors. A few visual-tactile policy learning methods showed that careful preprocessing of tactile signals, such as constructing visual-tactile point clouds \cite{34-huang20253d, VT-Refine} and extracting dense force maps from GelSight Mini images \cite{98-FARM}, allows standard vision encoders to be applied effectively to tactile signals. A larger body of works learns dedicated tactile encoder weights through a pretraining stage that precedes policy learning. Objectives used for pretraining include visual-tactile feature alignment \cite{78-george2025vital, 86-wu2025freetacman}, future tactile signal prediction \cite{65-xu2025exumi}, masked autoencoding \cite{56-zhu2025touch, 64-wu2025canonical, 46-zhao2025transferable}, and other self-supervised approaches \cite{38-guzey2023dexterity, 91-xu2025unit, 88-higuera2025sparsh, 102-Sparsh-X, zheng2026omnivtavisuotactileworldmodeling}. While low-dimensional force/torque signals are often encoded with simpler neural networks without pretraining, applying standard vision encoders to 2D spectrogram images converted from F/T signals has proven to be more effective in some cases \cite{67-ACP}, and self-supervised pretraining yields better results for tactile skin F/T readings distributed across a dexterous hand \cite{64-wu2025canonical}. 

\subsubsection{Refine vision-generated trajectories with reactive tactile policy} \label{4-1-2}
Modern imitation learning methods often predict and execute a sequence of future actions, often called an action chunk, in a single inference step \cite{ACT, chi2025diffusion}. This improves the temporal consistency of generated actions, with the cost that the robot operates in an open-loop manner within each action chunk, unable to react to events that occur before the next prediction. This differs from how humans use touch for contact-rich manipulation, where fast tactile feedback drives rapid response to contact events such as making and breaking contact, slip, and sudden change in force \cite{johansson2009coding}.
Inspired by human behaviors, a growing body of work pairs a vision-dominant policy that runs at the standard vision policy rate with a second, tactile-driven policy that runs at a higher frequency to make reactive adjustments between chunk predictions \cite{100-CR-DAgger, 27-xue2025reactive, zheng2026omnivtavisuotactileworldmodeling}.

A structural limitation of this hierarchical design is that the fast tactile policy’s high-frequency requirement constrains how complex its inputs and architecture can be. In practice, the fast policies operate with low-dimensional tactile signals such as F/T readings \cite{27-xue2025reactive, 100-CR-DAgger} or tactile marker displacements \cite{27-xue2025reactive, zheng2026omnivtavisuotactileworldmodeling}, rather than the high-resolution tactile imagery that many representation learning methods discussed in Section \ref{4-1-1} were designed to exploit.

\subsubsection{Adapt policy output for force-based control}
Aside from methods that incorporate tactile sensing into the policy's input, another line of work extends the policy's output. While recent imitation learning methods adopt position-based control and predict 6-DoF pose \cite{chi2025diffusion} or joint-angle targets \cite{ACT}, force-based control has long been the standard approach in classical robotics for handling contact-rich manipulation. By expanding the action space to include force targets alongside position targets, predicted actions can be executed through classical force-based controllers such as impedance and admittance control, allowing learned policies to inherit a mature body of control techniques rather than reinventing them \cite{villani2016force}. With an F/T sensor at the robot wrist, the policy can be trained to predict force-informed virtual position targets \cite{28-chen2025dexforce, 67-ACP, 101-UMI-FT, Comp-ACT}, or predict force targets directly alongside position targets \cite{71-ForceMimic, 99-zhou2025admittance, 100-CR-DAgger}. Similarly, when the gripper fingers are equipped with fingertip tactile sensors, the policy can be trained to predict target grasping force alongside target finger width to enable precise grasp control \cite{101-UMI-FT, 98-FARM}. 

Less settled is the tradeoff between compliant behaviors and how accurately a robot can follow the given position targets. Some works selected a fixed set of compliant controller gains suitable to solve all chosen tasks \cite{28-chen2025dexforce, 100-CR-DAgger}, while others find it necessary to have a different set of pre-set gains for each task \cite{Comp-ACT, 99-zhou2025admittance}. It is also common to apply compliant control when the robot is in contact with the environment and high-accuracy position control otherwise, conditioned on a manually set force threshold to determine whether contact is happening \cite{67-ACP, 71-ForceMimic, 98-FARM, 80-he2025foar}.

\subsubsection{Additional approaches}
Beyond the three patterns above, several other integration mechanisms have emerged. Drawing from the empirical observation that naive concatenation causes policies to ignore tactile feedback, one line of work manipulates the training process to force greater tactile reliance, randomly masking vision inputs throughout training \cite{47-romero2024eyesight} or blurring vision inputs at the early stages of training \cite{68-liu2025factr}. Recognizing that tactile signals become more informative after contact happens, another group of work found it beneficial to dynamically weight vision and tactile features based on the predicted likelihood of contact \cite{80-he2025foar, zheng2026omnivtavisuotactileworldmodeling}. A related work has explored policy composition rather than feature fusion, dynamically combining the score functions of separate vision and tactile experts \cite{90-chen2026multimodal}. A more recent direction uses tactile signals to steer a vision-only base policy at inference time \cite{89-zhang2026touchguide}.

\subsubsection{Discussion}
Tactile-augmented imitation learning methods have brought performance improvements to vision-only policies along several dimensions: appropriate force regulation \cite{27-xue2025reactive, 47-romero2024eyesight, 67-ACP, 71-ForceMimic, 80-he2025foar, 89-zhang2026touchguide}, reliable in-hand state estimation \cite{34-huang20253d, 65-xu2025exumi, 89-zhang2026touchguide, 90-chen2026multimodal}, robustness to visual occlusion  \cite{38-guzey2023dexterity, 56-zhu2025touch, 90-chen2026multimodal}, detection of contact events and transitions \cite{56-zhu2025touch, 89-zhang2026touchguide, 98-FARM}, and generalization to unseen objects \cite{38-guzey2023dexterity, 68-liu2025factr, 86-wu2025freetacman, 101-UMI-FT}. However, these gains are uneven across sensor types and integration strategies, motivating the discussion on tactile sensor heterogeneity in Section \ref{section:5-heterogeneity} and the open question below: 

\textbf{The functional roles of different tactile signal types in imitation learning remain an open question.} The available evidence suggests that while the gains in force regulation and contact event detection are largely sensor-agnostic \cite{27-xue2025reactive, 47-romero2024eyesight, 67-ACP, 71-ForceMimic, 80-he2025foar, 89-zhang2026touchguide, 56-zhu2025touch, 98-FARM}, in-hand state estimation and visual occlusion robustness tend to benefit from the geometric details available from high-resolution tactile sensors \cite{34-huang20253d, 65-xu2025exumi, 89-zhang2026touchguide, 90-chen2026multimodal, 38-guzey2023dexterity, 56-zhu2025touch}. However, it has also been reported that tactile resolution gains can saturate \cite{34-huang20253d}. These observations call for a more principled understanding of when richer tactile signals yield meaningful policy gains, and when simpler sensing suffices. 


\subsection{Tactile-Augmented Reinforcement Learning}\label{Sec:RL}
Reinforcement learning (RL) is a framework in which the robot learns a policy through trial-and-error interaction with its environment, optimizing its actions based on a reward signal that defines task success. For contact-rich manipulation, RL is well-suited to learning policies that react to high-frequency tactile signals and removes the need for hand-engineered contact strategies. The combination of tactile sensing and RL has a long history in the field \cite{chebotar2016self, van2015learning}, with tactile signals most frequently entering the RL pipeline as part of the policy's observations. This trial-and-error paradigm comes at a cost, however, as shaping an effective policy typically requires a large amount of interaction data. We discuss below how this challenge manifests for tactile-driven manipulation and how the field is addressing it, followed by a discussion of combining RL with IL, a growing direction in manipulation research, for tactile-driven manipulation. 

\subsubsection{Real-world RL}
Training reinforcement learning policies directly in the real world allows robots to acquire manipulation skills from the distribution of interactions they will encounter at deployment. This is appealing for tactile-based manipulation since obtaining accurate contact and rich tactile signals in simulation can be costly. Real-world tactile RL has been demonstrated both on tasks that require reacting to subtle changes in contact state, such as peg-in-hole insertion \cite{36-dong2021tactile, 31-lee2020making} and typing on a braille keyboard \cite{church2020deep}, and on tasks under challenging dynamic conditions, such as overhand throwing of deformable darts \cite{aslam2025dartbot}. 
Data efficiency is critical in this setting for several reasons: running episodes on real hardware is slow, automatic resets between episodes require careful engineering \cite{aslam2025dartbot, 36-dong2021tactile}, and tactile sensors are fragile and degrade under repeated contact. Choosing compact and task-appropriate tactile representations, such as marker flows extracted from the raw images of vision-based tactile sensors \cite{aslam2025dartbot, 36-dong2021tactile, church2020deep} or pretrained multimodal representations that fuse vision and force/torque signals \cite{31-lee2020making}, has proven to be an effective strategy. Appropriate RL design choices can also improve data efficiency for real-world tactile RL, including curriculum learning \cite{36-dong2021tactile}, stage-based dense rewards \cite{31-lee2020making}, and hindsight experience replay for sparse-reward settings \cite{church2020deep}. However, current real-world tactile RL is largely confined to small action spaces such as reduced end-effector degrees of freedom, limiting the range of tasks for which the approach is currently viable.

\subsubsection{Sim-to-real RL}
As simulation becomes faster and more physically accurate, a growing body of work first trains RL policies in simulation before deploying them on real hardware, bypassing the constraints of real-world RL.
Simulation also provides access to privileged information difficult to obtain in the real world, such as ground truth object pose and dynamics, enabling data-efficient methods such as asymmetric actor-critic and teacher-student distillation.

One line of work that has benefited the most from sim-to-real RL is dexterous in-hand manipulation, which often takes the form of rotating or translating objects in the robot hand through finger movements. OpenAI’s pivotal work \cite{OpenAI-dexterous} demonstrated the viability of sim-to-real RL for in-hand object reorientation, with real-world state estimation provided by a multi-camera setup. Adding tactile sensing to the system helps reduce such reliance on cameras \cite{49-pitz2023dextrous}. Subsequent works have demonstrated blind in-hand object rotation and translation \cite{khandate2023sampling, 63-yin2025learning, 32-yin2023rotating}, simultaneous in-hand object rotation and 3D shape reconstruction \cite{30-qi2023general}, and rotation under arbitrary palm orientations \cite{45-yang2025anyrotate}. These methods often choose to bypass complicated sim-to-real procedures for raw tactile signals by extracting simplified tactile features, such as binary contact \cite{32-yin2023rotating, khandate2023sampling, 63-yin2025learning} and discretized contact location \cite{30-qi2023general}. Even at reduced fidelity, including touch consistently improves generalization to unseen objects \cite{32-yin2023rotating, 30-qi2023general, 63-yin2025learning}.

In contrast, a growing body of recent works simulate and leverage richer and sometimes raw tactile sensor signals, including contact force \cite{83-noseworthy2025forge}, marker displacements \cite{chen2024general, zhao2024fots, 33-akinola2025tacsl}, and raw tactile images \cite{zhao2024fots, 33-akinola2025tacsl, church2022tactile}. Using advanced simulation and sim-to-real techniques described in Section \ref{Sec:Simulation}, these works have demonstrated successful zero-shot transfer of various types of insertion and nonprehensile tasks. So far, such results are primarily demonstrated on parallel-jaw grippers and similar systems.

\subsubsection{Combine RL and IL for tactile-driven manipulation}
Reinforcement learning and imitation learning have complementary strengths: RL can discover suitable strategies but suffers from sample inefficiency and costly real-world training, while imitation learning leverages demonstrations but struggles with hard-to-demonstrate behaviors. Combining them has thus become an active research area in robotic manipulation \cite{lei2025rl}.

Recent works show that a residual policy can be learned through real-world RL with dense rewards to provide small corrections to the base tactile-augmented imitation learning policy, improving overall policy performance \cite{MimicTouch, zhao2025touch}. Leveraging recent advances in tactile simulation \cite{33-akinola2025tacsl} and simulation-friendly sensors \cite{huang2026flexitac}, VT-Refine \cite{VT-Refine} demonstrates that tactile-augmented diffusion policies trained on real-world data \cite{34-huang20253d} can be fine-tuned directly in simulation using only a binary task-success reward. The fine-tuned policies transfer zero-shot to the real world and exhibit emergent tactile behaviors absent from the demonstrations, significantly outperforming the base policy on various bimanual assembly tasks. 
Another recent work \cite{sharpa-apple} proposes an imitation learning system augmented with RL-learned tactile primitives such as dexterous in-hand rotation and stable grasping. By delegating these complex low-level tactile skills to RL, the IL policy can focus on high-level planning, yielding significantly better results than end-to-end IL pipelines that attempt to learn such skills from scratch.

For tactile-driven manipulation, hybrid RL + IL approaches remain a much smaller body of work than standalone tactile RL or IL methods. This likely reflects that tactile RL + IL inherits many bottlenecks from tactile RL: real-world RL can degrade tactile sensors quickly, and sim-to-real transfer remains a delicate and complicated procedure for many tactile sensor types.

\subsubsection{Discussion}
The three threads discussed above highlight one of the central challenges of tactile-augmented reinforcement learning: the cost of data. Much of the progress in tactile-augmented RL can be viewed as attempts to improve the efficiency of acquiring, generating, and leveraging such data. Beyond these efforts, however, we believe another fundamental question remains largely unexplored: 

\textbf{Choosing tactile signals and representations for RL involves fundamental task-dependent tradeoffs between signal richness, data efficiency, and simulatability.}
Several considerations push toward compact, low-dimensional tactile features: they improve data efficiency for RL training and reduce sim-to-real gaps. Although simulation for complex tactile signals is maturing rapidly, whether richer signals actually yield meaningful RL gains remains an open question. Dexterous in-hand manipulation works have demonstrated impressive results with sparse or binary tactile signals, and the evidence on whether additional signal richness improves performance is mixed \cite{miller2026enhancing}. Some report that binary or low-dimensional contact signals are sufficient or even competitive with denser feedback \cite{khandate2023sampling, 32-yin2023rotating, 30-qi2023general}, while others find that dense-featured touch yields concrete gains for specific tasks, particularly when subtle contact information is needed to detect events such as incipient slip \cite{45-yang2025anyrotate}. Taken together, these discrepancies suggest that the benefit of richer tactile feedback is task-dependent rather than universal. The resulting tradeoffs among signal richness, data efficiency, and simulatability motivate further research into mapping task demands to suitable tactile features.

\subsection{Tactile-Augmented Vision-Language-Action Models}
\label{section:4-2}

Sections \ref{section:4-1} and \ref{Sec:RL} discussed learning paradigms that are typically trained from scratch on human demonstration or robot exploration data. Although these methods are applicable to a wide range of tasks, they require training a specific policy for each task. To reduce such burden, the broader robotic manipulation community has begun exploring a different paradigm: vision-language-action (VLA) models \cite{RT-2}. Trained with supervised objectives similar to those in Section \ref{section:4-1} but on larger-scale multi-task demonstration data, VLAs are built by equipping a pretrained vision language model (VLM), often called a \textit{backbone}, to output robot actions from visual observations and natural language task specifications. This mapping is often handled by a dedicated \textit{action expert}, which is a separate network that decodes the backbone's representations into robot control commands. Inheriting the rich representations and commonsense knowledge from the backbone pretrained on internet-scale data, VLAs are designed with the ultimate goal of generalizing across different tasks, objects, embodiments, environments, and language instructions through a single set of policy weights, with recent systems making encouraging progress in this direction \cite{pi0.5}. 

The improvements tactile sensing brings to vision-only imitation learning policies (Section \ref{section:4-1}) motivate analogous efforts in the VLA setting. However, this paradigm shift sets up new challenges for tactile sensing integration. In a policy trained from scratch, the policy's internal representations for both vision and touch are mostly shaped end-to-end by the task-specific datasets available. In a VLA, by contrast, the powerful backbone arrives with semantic priors learned from internet-scale image-text data with little notion of tactile or contact-level information. The architectural question is no longer how to fuse tactile with vision in a jointly trained policy, but where and how to engineer the tactile modality into a representation space that was trained without it. 
In contrast to Section \ref{section:4-1}, where high-resolution tactile imagery paired with representation learning has proven highly effective and remains a dominant approach, many tactile-augmented VLA methods opt to work with low-dimensional signals -- such as F/T and pressure-array readings -- encoded through linear projections or shallow multilayer perceptrons. Below, we further discuss the two major strategies for integrating tactile sensing into VLA pipelines. 

\subsubsection{Tactile as a co-equal input modality in the backbone} 
\label{section:4-3-1} One group of work treats the tactile modality 
equally as the pretrained backbone's native input modalities -- vision and language -- and adapts the backbone to accommodate them. Therefore, the technical question becomes how to reshape the backbone’s representation space without degrading its pretrained knowledge. One approach is adding a multimodal contrastive learning structure to the VLA training pipeline to maximize the mutual information shared between vision, language, and touch \cite{69-jones2025beyond, MLA}. Another approach is to directly use non-causal attention masks in the VLM backbone, allowing tactile input to fuse with other modalities, either at all times \cite{Tactile-VLA} or only during contact \cite{TacVLA}. This method shapes a deeply integrated representation space. Tactile-VLA \cite{Tactile-VLA} observes that the pretrained VLM's prior knowledge already includes a semantic understanding of physical interaction, and that connecting it to tactile signals via multimodal demonstrations is sufficient to unlock this understanding for downstream robot control.

\subsubsection{Tactile as add-on modules to the backbone} 
\label{section:4-2-2}
In contrast, another group of work takes the perspective that the pretrained VLM backbone's representation space is not only valuable, but also fragile under finetuning. Therefore, their design commitment is to separate the tactile stream from those of vision and language, preventing it from disrupting the backbone’s pretrained representation space. Directly contrasting the multimodal fusion-friendly attention mask design mentioned in Section \ref{section:4-3-1}, FD-VLA \cite{FD-VLA} designs directional attention masks in the VLM to limit fusion between tactile and VLM-native modalities. TA-VLA \cite{TA-VLA} provides a systematic exploration of where and how to introduce torque signals into a VLA, empirically finding that integrating torque into the VLA’s action expert rather than its VLM backbone yields the strongest results. Similar to the concept described in Section \ref{4-1-2}, T-Rex \cite{T-Rex} and AT-VLA \cite{AT-VLA} achieve rapid closed-loop reactions by injecting high-frequency tactile streams into the base VLA's action expert through cascaded flow matching and contact-gated adaptive cross-attention. Another method of tactile integration uses Mixture-of-Experts \cite{MoE} to fuse tactile features with the output of the backbone to form the input to the action expert \cite{ForceVLA, ForceVLA2}, or to compute residual corrections for the VLA’s raw action predictions \cite{sharpa-apple}. Building on the intuition that tactile perception contributes to manipulation on both the planning and control level, VLA-Touch \cite{VLA-Touch} uses a tactile-language model \cite{41-yu2024octopi} to help the VLM planner generate object property-informed instructions for a vision-only VLA, and uses a tactile-based interpolant controller to refine the raw actions predicted by the VLA. Recent work FTP-1 \cite{FTP-1} introduces a dedicated tactile expert parallel to the VLM and the action expert and pretrains it on \~3000 hours of data across 21 types of heterogeneous tactile sensors, offering a reusable initialization for downstream tactile policy training.

\subsubsection{Discussion} Tactile sensing improved the performance of vision-only VLA policies across diverse contact-rich tasks similar to those in Section \ref{section:4-1}. Although the two integration strategies discussed above make opposite architectural commitments, they share the following broader pattern when viewed across all works: 

\textbf{Existing tactile-augmented VLAs establish architectural feasibility, but substantial opportunities remain for realizing the VLA paradigm's core promise of generalization.} 
While a few works show that their tactile-augmented VLAs generalize across various novel objects \cite{ForceVLA} and language instructions \cite{69-jones2025beyond, Tactile-VLA}, most train one policy per task and evaluate it in the same narrow setting as the trained-from-scratch imitation learning policies in Section \ref{section:4-1}, with weak emphasis on generalization. 
More systematic evaluations in genuinely multi-task, language-conditioned, and novel-object settings would demonstrate the distinctive promises of the tactile-augmented VLA paradigm itself.

\subsection{Tactile Predictive Models for Manipulation}
\label{section:4-3}

A predictive model for manipulation takes the current state and a candidate action as input, and outputs the predicted future state. Such models support a predict-then-control paradigm: candidate actions are evaluated through their predicted outcomes, and the action most aligned with the task objective is selected. This idea underlies classical Model Predictive Control (MPC), which optimizes a finite-horizon action sequence against a dynamics model~\cite{richalet1978mph,mayne2000constrained}. \textit{World models}, a prominent paradigm emerging more broadly in robot learning, build on the same predict-then-control idea while reforming how the predictive model and its representations are built. 
Specifically, we distinguish two cases: a \textit{dynamics model} typically predicts transitions over a compact, task-specific state and is coupled to an external optimizer such as MPC, whereas a \textit{world model} extends the prediction target to high-dimensional observations or latent encodings learned by pretrained encoders. In both cases, action-conditioned prediction is what makes the model useful for control. These predictive models matter most in contact-rich tasks where success depends on how contact evolves -- slip, force change, making or breaking contact -- rather than on what the camera sees. Since that decisive future is local to the contact interface and often occluded, predicting it from touch is what informs control.
Consistent with this section's data-driven focus, the predictive models we cover are learned from interaction data rather than derived from analytical contact mechanics. We use \textit{tactile predictive models} as an umbrella term for this learned lineage, treating recent tactile world models as a scaling of predict-then-control to tactile or visual-tactile latents rather than a new paradigm.

\subsubsection{Predictive control with tactile dynamics models}
\label{subsubsec:deep-predictive-control}

Because contact is local and its mechanics are comparatively well understood, touch was a natural target for forward modeling. Initial efforts on tactile prediction model was largely developed as task-specific predictive-control pipelines rather than as general-purpose world models. Tian et al.~\cite{tian2019manipulationbyfeel} learned an action-conditioned model that forecasts future GelSight tactile images of a small object under candidate finger movement. They then embedded this predictor into sampling-based MPC to servo objects such as a ball, an analog stick, and a die toward a goal position specified directly as a raw tactile image.  Sutanto et al.~\cite{Sutanto_2019} followed a similar servoing paradigm, but represented tactile observations in a learned latent space rather than using raw tactile signals directly. 
LeTac-MPC~\cite{xu2024letacmpc} took a different route by embedding a differentiable MPC layer over a learned tactile representation and training the controller end-to-end for reactive grasping. This enabled stable grasping of objects with varying physical properties under external disturbance. 
Those predictive models are trained with interaction data on the same embodiment, either following an automatic exploration pattern or a human-defined trajectory. 
Together, these works established the appeal that the rest of this subsection inherits: contact is structured enough to model forward, and informative enough for closed-loop correction.

\subsubsection{Tactile world models for control}
\label{subsubsec:prediction-for-control}
World models are now a leading approach in vision-based control, where learning to predict how observations evolve under action has driven strong results on robotic manipulation~\cite{guo2025ctrl, wu2023daydreamer}. The visual-tactile world models below build on that success, extending visual world models to settings where contact carries the decisive signal. The works differ mainly in how the prediction is turned into an action: optimized over by an external planner, or consumed by a learned policy.

One line hands the forecast to an external action planner. RoboPack~\cite{ai2024robopack} learns a tactile-informed dynamics model that predicts future object states, using touch to infer latent physical properties that vision alone cannot reveal, and plans action with MPC.  VT-WM~\cite{higuera2026visuotactileworldmodels} carries the same scheme into pretrained latents for both modalities, then plans by sampling action sequences and rolling them through the predicted latents. A second line uses forecasts more directly inside the policy. OmniVTA~\cite{zheng2026omnivtavisuotactileworldmodeling} forecasts future vision and touch, and feeds the tactile forecast to a 60 Hz reflexive controller that corrects deviations between predicted and observed contact. By contrast, ViTacFormer~\cite{heng2025vitacformer} and DreamTacVLA~\cite{ye2026learningfeelfuturedreamtacvla} predict only the future tactile signal and condition action generation on that forecast. Recent works such as Dream-Tac \cite{dream-tac} attempt to harness the power of large pretrained video prediction models by fine-tuning them to predict future tactile readings and robot actions alongside robot camera views.

So far these methods work best on short-horizon, contact-reactive tasks where the decisive signal is immediately available, such as pushing~\cite{ai2024robopack, higuera2026visuotactileworldmodels}, wiping~\cite{higuera2026visuotactileworldmodels, zheng2026omnivtavisuotactileworldmodeling}, and stacking~\cite{higuera2026visuotactileworldmodels}, or correcting an action as soon as observed contact departs from the prediction~\cite{zheng2026omnivtavisuotactileworldmodeling, ye2026learningfeelfuturedreamtacvla}.  The longer-horizon prediction that originally motivated vision world models has not been demonstrated for touch, and whether it suits touch's inherently local nature is itself unclear.

\subsubsection{Tactile prediction as an auxiliary training objective}
\label{subsubsec:prediction-auxiliary}
 
A different strategy uses tactile prediction only during training.
An auxiliary loss that forecasts future tactile signals or contact is added to the policy-learning objective in the hope that the policy learns a more contact-aware representation. At deployment, the policy acts on current observations alone. This method applies to policies that are trained in different ways, including
imitation learning~\cite{niu2026learningversatilehumanoidmanipulation, 65-xu2025exumi}, reinforcement learning~\cite{31-lee2020making, miller2026enhancing}, and VLA policies~\cite{TA-VLA, MLA}, consistently reporting that the auxiliary tactile-prediction objective improves policy performance on contact-rich tasks over a no-prediction baseline, at no added inference cost.


\subsubsection{Discussion}
\label{subsubsec:wm-gaps}
 
While predict-then-control is a long-standing idea, its tactile and visual-tactile world-model instantiation is recent, and the optimal design choices are yet to be explored. Rather than offering firm conclusions, we close by raising several open questions that we view as crucial to advancing this direction:

\textbf{A tactile world model should be more than a high-fidelity signal predictor.} 
Many tactile predictive methods ~\cite{tian2019manipulationbyfeel, heng2025vitacformer, zheng2026omnivtavisuotactileworldmodeling} apply pixel- or token-level similarity metrics between predicted and ground-truth frames, which are inherited from the vision modality. But they might not be suitable to tactile modality. For touch, subtle spikes in contact may barely affect the averaged error, but are essential for task-relevant information.  We therefore believe that developing a more suitable metric for tactile signals will significantly improve the effectiveness of using touch in contact-rich manipulation. 


\textbf{Can foundation-scale prediction keep pace with contact dynamics?}
Robots need to quickly react to many tactile events, such as making and breaking contact, subtle changes in force, and incipient slip, whereas foundation-scale predictors ~\cite{higuera2026visuotactileworldmodels, zheng2026omnivtavisuotactileworldmodeling} run at much lower rates. 
It is essential for tactile world models to address this temporal mismatch between slow prediction and fast contact dynamics. We believe the solutions include developing either faster prediction models or mechanisms that explicitly separate slow predictive planning from fast reactive correction.
\section{OPEN PROBLEMS AND FUTURE DIRECTIONS}

The rapid development of tactile sensing, robot learning, and multimodal foundation models has created new opportunities for tactile-based manipulation. 
%
Yet the new tools and new possibilities introduce new challenges that
may play a critical role in determining the future of tactile-based manipulation:

\subsection{Tactile Sensor Heterogeneity and Transferability}
\label{section:5-heterogeneity}

As discussed in Section~\ref{Sec:Sensors}, tactile sensors exhibit substantial variation in signal formats and output characteristics. This heterogeneity has become a major obstacle to scaling tactile datasets and sharing knowledge across the community. In contrast to vision, where large datasets can often be reused across platforms, tactile datasets and models are frequently tied to specific sensors, making it difficult to transfer learned representations and policies across hardware.
A growing consensus is that addressing this challenge requires the development of \textbf{sensor-agnostic tactile representations}. One approach is learning shared latent representations directly from data collected from multiple vision-based tactile sensors \cite{46-zhao2025transferable,43-fenganytouch,88-higuera2025sparsh,92-sitr}, with recent work extending this effort to additionally include array- and state-based tactile sensors \cite{FTP-1}. Another uses physically meaningful quantities, such as contact geometry \cite{rodriguez2025cross} or contact force \cite{chen2026training}, as an intermediate representation shared across sensors. While promising progress has been made, developing representations that generalize across diverse tactile sensors and tasks remains an important open challenge.

\subsection{Making Tactile Sensing Lightweight}
Many manipulation tasks require robots to react rapidly to contact events, such as collisions, slip, or changes in contact conditions. As tactile sensors and algorithms become increasingly complex, the latency introduced by communication and computation can become a bottleneck for real-time manipulation. We therefore believe that making tactile sensing more lightweight remains an important research direction. Possible approaches include: (1) applying task-appropriate sensors rather than maximizing sensor signal richness; (2) developing low-dimensional tactile representations that support efficient perception and control \cite{27-xue2025reactive}; and (3) co-designing sensor hardware and algorithms, for example, by performing signal preprocessing and feature extraction directly on the sensor. Such approaches may improve the responsiveness and deployability of tactile-enabled robotic systems.

This also raises a broader question about \textbf{sensor design: what properties should future tactile sensors prioritize?} Despite the availability of rich tactile signals, many projects ended up using only low-dimensional tactile information. In the current practice, we believe sensor characteristics such as repeatability, durability, latency, and ease of integration are important for building an effective manipulation system. We therefore argue that, rather than pursuing increasingly complex designs and higher signal bandwidth alone, the community would benefit from robust, accessible, and reliable tactile sensors.

\subsection{How Should We Evaluate Tactile-driven Manipulation?} \label{sec:benchmark discussion}

For tactile-driven manipulation, a fundamental question remains: what does it mean to \emph{achieve general-purpose manipulation}?
Most existing work evaluates success rates on narrowly scoped tasks that are difficult to compare across studies and may not reflect broader manipulation requirements. As a result, it remains challenging to assess how much progress individual methods contribute toward general tactile-enabled manipulation.

Recent work has begun to consider broader evaluation criteria, including training data efficiency \cite{86-wu2025freetacman,miller2026enhancing}, task efficiency \cite{83-noseworthy2025forge,45-yang2025anyrotate}, interaction force \cite{83-noseworthy2025forge,98-FARM}, and generalization across objects \cite{68-liu2025factr,63-yin2025learning, ForceVLA}. These metrics better reflect the practical requirements of real-world manipulation. However, the community has yet to converge on a common suite of representative evaluation tasks. Such benchmarks would not only facilitate more meaningful comparison between methods, but also help align the trajectory of the field itself. Evaluation tasks shape the systems built around them; agreeing on representative tactile manipulation challenges would encourage researchers to develop methods that address shared and broadly meaningful goals rather than optimizing for isolated tasks. Recent benchmarking efforts in simulation-based tactile manipulation \cite{miller2026enhancing,luu2025manifeel,chen2026univtac} represent encouraging progress, and extending these efforts across a wider range of sensors and tasks remains an important direction for future research.

\subsection{Lessons from Mechanics-Driven to Data-Driven Methods} 


While recent progress in tactile-driven manipulation has been largely driven by data-driven approaches, the underlying physics of touch and contact-rich manipulation remain unchanged. Thus, insights accumulated from mechanics-driven methods continue to provide valuable structure for understanding contact interactions and designing effective manipulation systems. We believe that future advances will require not only scalable learning techniques but also a deeper understanding of the physical principles underlying touch. Such knowledge may further inform the design of more efficient tactile representations and learning architectures.

Some concepts established in mechanics-driven methods also remain underexplored in data-driven approaches. One example is \textbf{active touch}, where tactile perception is coupled with exploratory robot actions to acquire task-relevant information. As briefly discussed in Section \ref{sec:property perception}, active touch has primarily been studied for perception \cite{seminara2019active}. Estimates of object states and physical properties obtained by active exploration can also inform low-level manipulation strategies (Section \ref{sec:low-level control}). Such approaches are particularly useful when interacting with objects with unknown or challenging physical properties. 
Existing data-driven manipulation methods often do not include components for actively gaining object properties or other task-relevant information through exploration, but this capability can improve manipulation policies’ generalizability when handling new environments or novel objects. As the field moves toward manipulation in more diverse and complex physical environments, actively exploring the environment so as to gain task-relevant information may become increasingly important.



\section{CONCLUSION}

Tactile sensing provides robots with direct access to physical interactions at the contact interface, making it a unique and valuable modality for robotic manipulation. In this paper, we reviewed the development of tactile-based manipulation from both mechanics-driven and data-driven perspectives. Viewed through the history of the field, tactile manipulation has evolved from explicit reasoning about contact phenomena toward learning tactile representations and manipulation policies directly from data. Despite their methodological differences, both paradigms seek to answer the same fundamental question: how can robots use touch to better understand and interact with the physical world? We hope this review provides a useful reference for researchers working toward that goal.

\bibliographystyle{ar-style3}
\bibliography{ref}

@ARTICLE{dahiya2009tactile,
    author={Dahiya, Ravinder S. and Metta, Giorgio and Valle, Maurizio and Sandini, Giulio},
    journal={IEEE Transactions on Robotics}, 
    title={Tactile Sensing—From Humans to Humanoids}, 
    year={2010},
    volume={26},
    number={1},
    pages={1-20},
    doi={10.1109/TRO.2009.2033627}
}

@Article{yuan2017gelsight,
    AUTHOR = {Yuan, Wenzhen and Dong, Siyuan and Adelson, Edward H.},
    TITLE = {GelSight: High-Resolution Robot Tactile Sensors for Estimating Geometry and Force},
    JOURNAL = {Sensors},
    VOLUME = {17},
    YEAR = {2017},
    NUMBER = {12},
    ARTICLE-NUMBER = {2762},
    URL = {https://www.mdpi.com/1424-8220/17/12/2762},
    PubMedID = {29186053},
    ISSN = {1424-8220},
    DOI = {10.3390/s17122762}
}

@article{schneiter1984optical,
  title={An optical tactile sensor for manipulators},
  author={Schneiter, John L and Sheridan, Thomas B},
  journal={Robotics and computer-integrated manufacturing},
  volume={1},
  number={1},
  pages={65--71},
  year={1984},
  publisher={Elsevier}
}

@article{ward2018tactip,
  title={The TacTip Family: Soft Optical Tactile Sensors with 3D-Printed Biomimetic Morphologies},
  author={Ward-Cherrier, Benjamin and Pestell, Nicholas and Cramphorn, Luke and Winstone, Benjamin and Giannaccini, Maria Elena and Rossiter, Jonathan and Lepora, Nathan F},
  journal={Soft robotics},
  volume={5},
  number={2},
  pages={216--227},
  year={2018},
  publisher={SAGE Publications Sage CA: Los Angeles, CA}
}

@INPROCEEDINGS{lin2023dtact,
  author={Lin, Changyi and Lin, Ziqi and Wang, Shaoxiong and Xu, Huazhe},
  booktitle={2023 IEEE International Conference on Robotics and Automation (ICRA)}, 
  title={DTact: A Vision-Based Tactile Sensor that Measures High-Resolution 3D Geometry Directly from Darkness}, 
  year={2023},
  volume={},
  number={},
  pages={10359-10366},
  doi={10.1109/ICRA48891.2023.10160796}
}

@ARTICLE{tomo2018new,
  author={Tomo, Tito Pradhono and Regoli, Massimo and Schmitz, Alexander and Natale, Lorenzo and Kristanto, Harris and Somlor, Sophon and Jamone, Lorenzo and Metta, Giorgio and Sugano, Shigeki},
  journal={IEEE Robotics and Automation Letters}, 
  title={A New Silicone Structure for uSkin—A Soft, Distributed, Digital 3-Axis Skin Sensor and Its Integration on the Humanoid Robot iCub}, 
  year={2018},
  volume={3},
  number={3},
  pages={2584-2591},
  doi={10.1109/LRA.2018.2812915}
}

@article{yan2021soft,
    author = {Youcan Yan  and Zhe Hu  and Zhengbao Yang  and Wenzhen Yuan  and Chaoyang Song  and Jia Pan  and Yajing Shen },
    title = {Soft magnetic skin for super-resolution tactile sensing with force self-decoupling},
    journal = {Science Robotics},
    volume = {6},
    number = {51},
    pages = {eabc8801},
    year = {2021},
    doi = {10.1126/scirobotics.abc8801},
    URL = {https://www.science.org/doi/abs/10.1126/scirobotics.abc8801},
    eprint = {https://www.science.org/doi/pdf/10.1126/scirobotics.abc8801},
}

@ARTICLE{gomes2021generation,
  author={Gomes, Daniel Fernandes and Paoletti, Paolo and Luo, Shan},
  journal={IEEE Robotics and Automation Letters}, 
  title={Generation of GelSight Tactile Images for Sim2Real Learning}, 
  year={2021},
  volume={6},
  number={2},
  pages={4177-4184},
  doi={10.1109/LRA.2021.3063925}
}

@ARTICLE{si2022taxim,
  author={Si, Zilin and Yuan, Wenzhen},
  journal={IEEE Robotics and Automation Letters}, 
  title={Taxim: An Example-Based Simulation Model for GelSight Tactile Sensors}, 
  year={2022},
  volume={7},
  number={2},
  pages={2361-2368},
  doi={10.1109/LRA.2022.3142412}
}

@ARTICLE{wang2022tacto,
  author={Wang, Shaoxiong and Lambeta, Mike and Chou, Po-Wei and Calandra, Roberto},
  journal={IEEE Robotics and Automation Letters}, 
  title={TACTO: A Fast, Flexible, and Open-Source Simulator for High-Resolution Vision-Based Tactile Sensors}, 
  year={2022},
  volume={7},
  number={2},
  pages={3930-3937},
  doi={10.1109/LRA.2022.3146945}
}

@INPROCEEDINGS{ding2020sim,
  author={Ding, Zihan and Lepora, Nathan F. and Johns, Edward},
  booktitle={2020 IEEE International Conference on Robotics and Automation (ICRA)}, 
  title={Sim-to-Real Transfer for Optical Tactile Sensing}, 
  year={2020},
  volume={},
  number={},
  pages={1639-1645},
  doi={10.1109/ICRA40945.2020.9197512}
}

@INPROCEEDINGS{agarwal2021simulation,
  author={Agarwal, Arpit and Man, Timothy and Yuan, Wenzhen},
  booktitle={2021 IEEE International Conference on Robotics and Automation (ICRA)}, 
  title={Simulation of Vision-based Tactile Sensors using Physics based Rendering}, 
  year={2021},
  volume={},
  number={},
  pages={1-7},
  doi={10.1109/ICRA48506.2021.9561122}
}

@INPROCEEDINGS{jianu2022reducing,
  author={Jianu, Tudor and Gomes, Daniel Fernandes and Luo, Shan},
  booktitle={2022 International Conference on Robotics and Automation (ICRA)}, 
  title={Reducing Tactile Sim2Real Domain Gaps via Deep Texture Generation Networks}, 
  year={2022},
  volume={},
  number={},
  pages={8305-8311},
  doi={10.1109/ICRA46639.2022.9811801}
}

@article{zhao2023skill,
    title = {Skill generalization of tubular object manipulation with tactile sensing and Sim2Real learning},
    journal = {Robotics and Autonomous Systems},
    volume = {160},
    pages = {104321},
    year = {2023},
    issn = {0921-8890},
    doi = {https://doi.org/10.1016/j.robot.2022.104321},
    url = {https://www.sciencedirect.com/science/article/pii/S092188902200210X},
    author = {Yongqiang Zhao and Xingshuo Jing and Kun Qian and Daniel Fernandes Gomes and Shan Luo},
}

@INPROCEEDINGS{pezzementi2010characterization,
  author={Pezzementi, Zachary and Jantho, Erica and Estrade, Lucas and Hager, Gregory D.},
  booktitle={2010 IEEE Haptics Symposium}, 
  title={Characterization and simulation of tactile sensors}, 
  year={2010},
  volume={},
  number={},
  pages={199-205},
  doi={10.1109/HAPTIC.2010.5444654}
}

@INPROCEEDINGS{narang2021sim,
  author={Narang, Yashraj and Sundaralingam, Balakumar and Macklin, Miles and Mousavian, Arsalan and Fox, Dieter},
  booktitle={2021 IEEE International Conference on Robotics and Automation (ICRA)}, 
  title={Sim-to-Real for Robotic Tactile Sensing via Physics-Based Simulation and Learned Latent Projections}, 
  year={2021},
  volume={},
  number={},
  pages={6444-6451},
  doi={10.1109/ICRA48506.2021.9561969}
}

@InProceedings{ruppel2019simulation,
    author="Ruppel, Philipp
    and Jonetzko, Yannick
    and G{\"o}rner, Michael
    and Hendrich, Norman
    and Zhang, Jianwei",
    editor="Strand, Marcus
    and Dillmann, R{\"u}diger
    and Menegatti, Emanuele
    and Ghidoni, Stefano",
    title="Simulation of the SynTouch BioTac Sensor",
    booktitle="Intelligent Autonomous Systems 15",
    year="2019",
    publisher="Springer International Publishing",
    address="Cham",
    pages="374--387",
    isbn="978-3-030-01370-7"
}

@INPROCEEDINGS{suresh2022shapemap,
  author={Suresh, Sudharshan and Si, Zilin and Mangelson, Joshua G. and Yuan, Wenzhen and Kaess, Michael},
  booktitle={2022 International Conference on Robotics and Automation (ICRA)}, 
  title={ShapeMap 3-D: Efficient shape mapping through dense touch and vision}, 
  year={2022},
  volume={},
  number={},
  pages={7073-7080},
  doi={10.1109/ICRA46639.2022.9812040}
}

@INPROCEEDINGS{si2022grasp,
  author={Si, Zilin and Zhu, Zirui and Agarwal, Arpit and Anderson, Stuart and Yuan, Wenzhen},
  booktitle={2022 IEEE/RSJ International Conference on Intelligent Robots and Systems (IROS)}, 
  title={Grasp Stability Prediction with Sim-to-Real Transfer from Tactile Sensing}, 
  year={2022},
  volume={},
  number={},
  pages={7809-7816},
  doi={10.1109/IROS47612.2022.9981863}
}

@INPROCEEDINGS{sferrazza2020learning,
  author={Sferrazza, Carmelo and Bi, Thomas and D’Andrea, Raffaello},
  booktitle={2020 IEEE/RSJ International Conference on Intelligent Robots and Systems (IROS)}, 
  title={Learning the sense of touch in simulation: a sim-to-real strategy for vision-based tactile sensing}, 
  year={2020},
  volume={},
  number={},
  pages={4389-4396},
  doi={10.1109/IROS45743.2020.9341285}
}

@ARTICLE{chen2024general,
  author={Chen, Weihang and Xu, Jing and Xiang, Fanbo and Yuan, Xiaodi and Su, Hao and Chen, Rui},
  journal={IEEE Transactions on Robotics}, 
  title={General-Purpose Sim2Real Protocol for Learning Contact-Rich Manipulation With Marker-Based Visuotactile Sensors}, 
  year={2024},
  volume={40},
  number={},
  pages={1509-1526},
  doi={10.1109/TRO.2024.3352969}
}

@InProceedings{church2022tactile,
  title = 	 {Tactile Sim-to-Real Policy Transfer via Real-to-Sim Image Translation},
  author =       {Church, Alex and Lloyd, John and hadsell, raia and Lepora, Nathan F.},
  booktitle = 	 {Proceedings of the 5th Conference on Robot Learning},
  pages = 	 {1645--1654},
  year = 	 {2022},
  editor = 	 {Faust, Aleksandra and Hsu, David and Neumann, Gerhard},
  volume = 	 {164},
  series = 	 {Proceedings of Machine Learning Research},
  month = 	 {08--11 Nov},
  publisher =    {PMLR},
  url = 	 {https://proceedings.mlr.press/v164/church22a.html},
}

@ARTICLE{lin2022tactile,
  author={Lin, Yijiong and Lloyd, John and Church, Alex and Lepora, Nathan F.},
  journal={IEEE Robotics and Automation Letters}, 
  title={Tactile Gym 2.0: Sim-to-Real Deep Reinforcement Learning for Comparing Low-Cost High-Resolution Robot Touch}, 
  year={2022},
  volume={7},
  number={4},
  pages={10754-10761},
  doi={10.1109/LRA.2022.3195195}
}

@inproceedings{dang2026hydroshear,
    author = {Dang, An and Lee, Jayjun and Mukadam, Mustafa and Wu, X Alice and Bucher, Bernadette and Nambi, Manikantan and Fazeli, Nima},
    title = {{Hydroshear: Hydroelastic shear simulation for tactile sim-to-real reinforcement learning}},
    booktitle = {Proceedings of Robotics: Science and Systems},
    year = {2026},
}

@INPROCEEDINGS{Eguiluz2017Handover,
  author={Eguíluz, A. Gómez and Rañó, I. and Coleman, S. A. and McGinnity, T. M.},
  booktitle={2017 IEEE International Conference on Robotics and Automation (ICRA)}, 
  title={Reliable object handover through tactile force sensing and effort control in the Shadow Robot hand}, 
  year={2017},
  volume={},
  number={},
  pages={372-377},
  doi={10.1109/ICRA.2017.7989048}}

@INPROCEEDINGS{Hou2020SharedGrasping, 
    AUTHOR    = {Yifan Hou AND Zhenzhong Jia AND Matthew Mason}, 
    TITLE     = {{Manipulation with Shared Grasping}}, 
    BOOKTITLE = {Proceedings of Robotics: Science and Systems}, 
    YEAR      = {2020}, 
    ADDRESS   = {Corvalis, Oregon, USA}, 
    MONTH     = {July}, 
    DOI       = {10.15607/RSS.2020.XVI.086} 
}

@INPROCEEDINGS{Oller2024ContactMode, 
    AUTHOR    = {Miquel Oller AND Dmitry Berenson AND Nima Fazeli}, 
    TITLE     = {{Tactile-Driven Non-Prehensile Object Manipulation via Extrinsic Contact Mode Control}}, 
    BOOKTITLE = {Proceedings of Robotics: Science and Systems}, 
    YEAR      = {2024}, 
    ADDRESS   = {Delft, Netherlands}, 
    MONTH     = {July}, 
    DOI       = {10.15607/RSS.2024.XX.135} 
}

@INPROCEEDINGS{dean2019whole,
  author={Dean-Leon, Emmanuel and Guadarrama-Olvera, J. Rogelio and Bergner, Florian and Cheng, Gordon},
  booktitle={2019 International Conference on Robotics and Automation (ICRA)}, 
  title={Whole-Body Active Compliance Control for Humanoid Robots with Robot Skin}, 
  year={2019},
  volume={},
  number={},
  pages={5404-5410},
  doi={10.1109/ICRA.2019.8793258}
}

@ARTICLE{huang2024normalflow,
  author={Huang, Hung-Jui and Kaess, Michael and Yuan, Wenzhen},
  journal={IEEE Robotics and Automation Letters}, 
  title={NormalFlow: Fast, Robust, and Accurate Contact-Based Object 6DoF Pose Tracking With Vision-Based Tactile Sensors}, 
  year={2025},
  volume={10},
  number={1},
  pages={452-459},
  doi={10.1109/LRA.2024.3505815}
}

@article{bauza2023tac2pose,
    author = {Maria Bauza and Antonia Bronars and Alberto Rodriguez},
    title ={Tac2Pose: Tactile object pose estimation from the first touch},
    journal = {The International Journal of Robotics Research},
    volume = {42},
    number = {13},
    pages = {1185-1209},
    year = {2023},
    doi = {10.1177/02783649231196925},
}

@article{bauza2024Simple,
author = {Maria Bauza  and Antonia Bronars  and Yifan Hou  and Ian Taylor  and Nikhil Chavan-Dafle  and Alberto Rodriguez },
title = {SimPLE, a visuotactile method learned in simulation to precisely pick, localize, regrasp, and place objects},
journal = {Science Robotics},
volume = {9},
number = {91},
pages = {eadi8808},
year = {2024},
doi = {10.1126/scirobotics.adi8808},
URL = {https://www.science.org/doi/abs/10.1126/scirobotics.adi8808},
eprint = {https://www.science.org/doi/pdf/10.1126/scirobotics.adi8808},
}

@INPROCEEDINGS{pai2023tactofind,
  author={Pai, Sameer and Chen, Tao and Tippur, Megha and Adelson, Edward and Gupta, Abhishek and Agrawal, Pulkit},
  booktitle={2023 IEEE International Conference on Robotics and Automation (ICRA)}, 
  title={TactoFind: A Tactile Only System for Object Retrieval}, 
  year={2023},
  volume={},
  number={},
  pages={8025-8032},
  doi={10.1109/ICRA48891.2023.10160289}
}

@INPROCEEDINGS{li2014localization,
  author={Li, Rui and Platt, Robert and Yuan, Wenzhen and ten Pas, Andreas and Roscup, Nathan and Srinivasan, Mandayam A. and Adelson, Edward},
  booktitle={2014 IEEE/RSJ International Conference on Intelligent Robots and Systems}, 
  title={Localization and manipulation of small parts using GelSight tactile sensing}, 
  year={2014},
  volume={},
  number={},
  pages={3988-3993},
  doi={10.1109/IROS.2014.6943123}
}

@ARTICLE{veiga2018grip,
  author={Veiga, Filipe and Peters, Jan and Hermans, Tucker},
  journal={IEEE Transactions on Haptics}, 
  title={Grip Stabilization of Novel Objects Using Slip Prediction}, 
  year={2018},
  volume={11},
  number={4},
  pages={531-542},
  doi={10.1109/TOH.2018.2837744}
}

@INPROCEEDINGS{howe1989sensing,
  author={Howe, R.D. and Cutkosky, M.R.},
  booktitle={Proceedings, 1989 International Conference on Robotics and Automation}, 
  title={Sensing skin acceleration for slip and texture perception}, 
  year={1989},
  volume={},
  number={},
  pages={145-150 vol.1},
  doi={10.1109/ROBOT.1989.99981}
}

@INPROCEEDINGS{yamada1994tactile,
  author={Yamada, Y. and Cutkosky, M.R.},
  booktitle={Proceedings of the 1994 IEEE International Conference on Robotics and Automation}, 
  title={Tactile sensor with 3-axis force and vibration sensing functions and its application to detect rotational slip}, 
  year={1994},
  volume={},
  number={},
  pages={3550-3557 vol.4},
  doi={10.1109/ROBOT.1994.351572}
}

@ARTICLE{melchiorri2002slip,
  author={Melchiorri, C.},
  journal={IEEE/ASME Transactions on Mechatronics}, 
  title={Slip detection and control using tactile and force sensors}, 
  year={2000},
  volume={5},
  number={3},
  pages={235-243},
  doi={10.1109/3516.868914}
}

@INPROCEEDINGS{ikeda2004grip,
  author={Ikeda, A. and Kurita, Y. and Ueda, J. and Matsumoto, Y. and Ogasawara, T.},
  booktitle={2004 IEEE/RSJ International Conference on Intelligent Robots and Systems (IROS) (IEEE Cat. No.04CH37566)}, 
  title={Grip force control for an elastic finger using vision-based incipient slip feedback}, 
  year={2004},
  volume={1},
  number={},
  pages={810-815 vol.1},
  doi={10.1109/IROS.2004.1389452}
}

@INPROCEEDINGS{yuan2015measurement,
  author={Yuan, Wenzhen and Li, Rui and Srinivasan, Mandayam A. and Adelson, Edward H.},
  booktitle={2015 IEEE International Conference on Robotics and Automation (ICRA)}, 
  title={Measurement of shear and slip with a GelSight tactile sensor}, 
  year={2015},
  volume={},
  number={},
  pages={304-311},
  doi={10.1109/ICRA.2015.7139016}
}

@INPROCEEDINGS{reinecke2014experimental,
  author={Reinecke, Jens and Dietrich, Alexander and Schmidt, Florian and Chalon, Maxime},
  booktitle={2014 IEEE International Conference on Robotics and Automation (ICRA)}, 
  title={Experimental comparison of slip detection strategies by tactile sensing with the BioTac® on the DLR hand arm system}, 
  year={2014},
  volume={},
  number={},
  pages={2742-2748},
  doi={10.1109/ICRA.2014.6907252}
}

@INPROCEEDINGS{su2015force,
  author={Su, Zhe and Hausman, Karol and Chebotar, Yevgen and Molchanov, Artem and Loeb, Gerald E. and Sukhatme, Gaurav S. and Schaal, Stefan},
  booktitle={2015 IEEE-RAS 15th International Conference on Humanoid Robots (Humanoids)}, 
  title={Force estimation and slip detection/classification for grip control using a biomimetic tactile sensor}, 
  year={2015},
  volume={},
  number={},
  pages={297-303},
  doi={10.1109/HUMANOIDS.2015.7363558}
}

@InProceedings{meier2016tactile,
    author="Meier, Martin
    and Patzelt, Florian
    and Haschke, Robert
    and Ritter, Helge J.",
    editor="Villa, Alessandro E.P.
    and Masulli, Paolo
    and Pons Rivero, Antonio Javier",
    title="Tactile Convolutional Networks for Online Slip and Rotation Detection",
    booktitle="Artificial Neural Networks and Machine Learning -- ICANN 2016",
    year="2016",
    publisher="Springer International Publishing",
    address="Cham",
    pages="12--19",
    isbn="978-3-319-44781-0"
}

@ARTICLE{james2018slip,
  author={James, Jasper Wollaston and Pestell, Nicholas and Lepora, Nathan F.},
  journal={IEEE Robotics and Automation Letters}, 
  title={Slip Detection With a Biomimetic Tactile Sensor}, 
  year={2018},
  volume={3},
  number={4},
  pages={3340-3346},
  doi={10.1109/LRA.2018.2852797}
}

@ARTICLE{sui2021incipient,
  author={Sui, Ruomin and Zhang, Lunwei and Li, Tiemin and Jiang, Yao},
  journal={IEEE Sensors Journal}, 
  title={Incipient Slip Detection Method With Vision-Based Tactile Sensor Based on Distribution Force and Deformation}, 
  year={2021},
  volume={21},
  number={22},
  pages={25973-25985},
  doi={10.1109/JSEN.2021.3119060}
}

@article{khamis2018papillarray,
    title = {PapillArray: An incipient slip sensor for dexterous robotic or prosthetic manipulation – design and prototype validation},
    journal = {Sensors and Actuators A: Physical},
    volume = {270},
    pages = {195-204},
    year = {2018},
    issn = {0924-4247},
    doi = {https://doi.org/10.1016/j.sna.2017.12.058},
    url = {https://www.sciencedirect.com/science/article/pii/S0924424717313419},
    author = {Heba Khamis and Raquel {Izquierdo Albero} and Matteo Salerno and Ahmad {Shah Idil} and Andrew Loizou and Stephen J. Redmond},
}

@INPROCEEDINGS{karayiannidis2014online,
  author={Karayiannidis, Yiannis and Smith, Christian and Viña, Francisco E. and Kragic, Danica},
  booktitle={2014 IEEE International Conference on Robotics and Automation (ICRA)}, 
  title={Online contact point estimation for uncalibrated tool use}, 
  year={2014},
  volume={},
  number={},
  pages={2488-2494},
  doi={10.1109/ICRA.2014.6907206}
}

@INPROCEEDINGS{molchanov2016contact,
  author={Molchanov, Artem and Kroemer, Oliver and Su, Zhe and Sukhatme, Gaurav S.},
  booktitle={2016 IEEE/RSJ International Conference on Intelligent Robots and Systems (IROS)}, 
  title={Contact localization on grasped objects using tactile sensing}, 
  year={2016},
  volume={},
  number={},
  pages={216-222},
  doi={10.1109/IROS.2016.7759058}
}

@INPROCEEDINGS{yu2016more,
  author={Yu, Kuan-Ting and Bauza, Maria and Fazeli, Nima and Rodriguez, Alberto},
  booktitle={2016 IEEE/RSJ International Conference on Intelligent Robots and Systems (IROS)}, 
  title={More than a million ways to be pushed. A high-fidelity experimental dataset of planar pushing}, 
  year={2016},
  volume={},
  number={},
  pages={30-37},
  doi={10.1109/IROS.2016.7758091}
}

@INPROCEEDINGS{atkeson1985rigid,
  author={Atkeson, Chirstopher G. and An, Chae H. and Hollerbach, John M.},
  booktitle={1985 24th IEEE Conference on Decision and Control}, 
  title={Rigid body load identification for manipulators}, 
  year={1985},
  volume={},
  number={},
  pages={996-1002},
  doi={10.1109/CDC.1985.268649}
}

@INPROCEEDINGS{zhang2019leveraging,
  author={Zhang, Kevin and Sharma, Mohit and Veloso, Manuela and Kroemer, Oliver},
  booktitle={2019 IEEE-RAS 19th International Conference on Humanoid Robots (Humanoids)}, 
  title={Leveraging Multimodal Haptic Sensory Data for Robust Cutting}, 
  year={2019},
  volume={},
  number={},
  pages={409-416},
  doi={10.1109/Humanoids43949.2019.9035073}
}

@INPROCEEDINGS{Li-tactile-servoing, 
    AUTHOR    = {Qiang Li AND Carsten SchÃ¼rmann AND Robert Haschke AND Helge Ritter}, 
    TITLE     = {A Control Framework for Tactile Servoing}, 
    BOOKTITLE = {Proceedings of Robotics: Science and Systems}, 
    YEAR      = {2013}, 
    ADDRESS   = {Berlin, Germany}, 
    MONTH     = {June},
    DOI       = {10.15607/RSS.2013.IX.045} 
}

@article{Jain2013ClutterWhole,
    author = {Jain, Advait and Killpack, Marc D and Edsinger, Aaron and Kemp, Charles C},
    title = {Reaching in clutter with whole-arm tactile sensing},
    year = {2013},
    issue_date = {Apr 2013},
    publisher = {Sage Publications, Inc.},
    address = {USA},
    volume = {32},
    number = {4},
    issn = {0278-3649},
    url = {https://doi.org/10.1177/0278364912471865},
    doi = {10.1177/0278364912471865},
    journal = {Int. J. Rob. Res.},
    month = apr,
    pages = {458–482},
    numpages = {25}
}

@INPROCEEDINGS{li2015visuo,
  author={Li, Qiang and Haschke, Robert and Ritter, Helge},
  booktitle={2015 IEEE-RAS 15th International Conference on Humanoid Robots (Humanoids)}, 
  title={A visuo-tactile control framework for manipulation and exploration of unknown objects}, 
  year={2015},
  volume={},
  number={},
  pages={610-615},
  doi={10.1109/HUMANOIDS.2015.7363434}
}

@ARTICLE{VanWyk2018PegInHole,
  author={Van Wyk, Karl and Culleton, Mark and Falco, Joe and Kelly, Kevin},
  journal={IEEE Transactions on Robotics}, 
  title={Comparative Peg-in-Hole Testing of a Force-Based Manipulation Controlled Robotic Hand}, 
  year={2018},
  volume={34},
  number={2},
  pages={542-549},
  doi={10.1109/TRO.2018.2791591}
}

@INPROCEEDINGS{li2017human,
  author={Li, Xiaoqing and Li, Rui and Qiao, Hong and Ma, Chao and Li, Liang},
  booktitle={2017 IEEE/RSJ International Conference on Intelligent Robots and Systems (IROS)}, 
  title={Human-inspired compliant strategy for peg-in-hole assembly using environmental constraint and coarse force information}, 
  year={2017},
  volume={},
  number={},
  pages={4743-4748},
  doi={10.1109/IROS.2017.8206348}
}

@ARTICLE{lepora2017exploratory,
  author={Lepora, Nathan F. and Aquilina, Kirsty and Cramphorn, Luke},
  journal={IEEE Robotics and Automation Letters}, 
  title={Exploratory Tactile Servoing With Active Touch}, 
  year={2017},
  volume={2},
  number={2},
  pages={1156-1163},
  doi={10.1109/LRA.2017.2662071}
}

@ARTICLE{lepora2019pixels,
  author={Lepora, Nathan F. and Church, Alex and de Kerckhove, Conrad and Hadsell, Raia and Lloyd, John},
  journal={IEEE Robotics and Automation Letters}, 
  title={From Pixels to Percepts: Highly Robust Edge Perception and Contour Following Using Deep Learning and an Optical Biomimetic Tactile Sensor}, 
  year={2019},
  volume={4},
  number={2},
  pages={2101-2107},
  doi={10.1109/LRA.2019.2899192}
}

@ARTICLE{romano2011human,
  author={Romano, Joseph M. and Hsiao, Kaijen and Niemeyer, Günter and Chitta, Sachin and Kuchenbecker, Katherine J.},
  journal={IEEE Transactions on Robotics}, 
  title={Human-Inspired Robotic Grasp Control With Tactile Sensing}, 
  year={2011},
  volume={27},
  number={6},
  pages={1067-1079},
  doi={10.1109/TRO.2011.2162271}
}

@Article{veiga2020grip,
    AUTHOR = {Veiga, Filipe and Edin, Benoni and Peters, Jan},
    TITLE = {Grip Stabilization through Independent Finger Tactile Feedback Control},
    JOURNAL = {Sensors},
    VOLUME = {20},
    YEAR = {2020},
    NUMBER = {6},
    ARTICLE-NUMBER = {1748},
    URL = {https://www.mdpi.com/1424-8220/20/6/1748},
    PubMedID = {32245193},
    ISSN = {1424-8220},
    DOI = {10.3390/s20061748}
}

@ARTICLE{calandra2018more,
  author={Calandra, Roberto and Owens, Andrew and Jayaraman, Dinesh and Lin, Justin and Yuan, Wenzhen and Malik, Jitendra and Adelson, Edward H. and Levine, Sergey},
  journal={IEEE Robotics and Automation Letters}, 
  title={More Than a Feeling: Learning to Grasp and Regrasp Using Vision and Touch}, 
  year={2018},
  volume={3},
  number={4},
  pages={3300-3307},
  doi={10.1109/LRA.2018.2852779}
}

@INPROCEEDINGS{hogan2018tactile,
  author={Hogan, Francois R. and Bauza, Maria and Canal, Oleguer and Donlon, Elliott and Rodriguez, Alberto},
  booktitle={2018 IEEE/RSJ International Conference on Intelligent Robots and Systems (IROS)}, 
  title={Tactile Regrasp: Grasp Adjustments via Simulated Tactile Transformations}, 
  year={2018},
  volume={},
  number={},
  pages={2963-2970},
  doi={10.1109/IROS.2018.8593528}
}

@INPROCEEDINGS{kolamuri2021improving,
  author={Kolamuri, Raj and Si, Zilin and Zhang, Yufan and Agarwal, Arpit and Yuan, Wenzhen},
  booktitle={2021 IEEE/RSJ International Conference on Intelligent Robots and Systems (IROS)}, 
  title={Improving Grasp Stability with Rotation Measurement from Tactile Sensing}, 
  year={2021},
  volume={},
  number={},
  pages={6809-6816},
  doi={10.1109/IROS51168.2021.9636488}
}

@INPROCEEDINGS{ajoudani2016reflex,
  author={Ajoudani, Arash and Hocaoglu, Elif and Altobelli, Alessandro and Rossi, Matteo and Battaglia, Edoardo and Tsagarakis, Nikos and Bicchi, Antonio},
  booktitle={2016 IEEE International Conference on Robotics and Automation (ICRA)}, 
  title={Reflex control of the Pisa/IIT SoftHand during object slippage}, 
  year={2016},
  volume={},
  number={},
  pages={1972-1979},
  doi={10.1109/ICRA.2016.7487344}
}

@ARTICLE{zheng2022autonomous,
  author={Zheng, Yi and Veiga, Filipe Fernandes and Peters, Jan and Santos, Veronica J.},
  journal={IEEE Transactions on Robotics}, 
  title={Autonomous Learning of Page Flipping Movements via Tactile Feedback}, 
  year={2022},
  volume={38},
  number={5},
  pages={2734-2749},
  doi={10.1109/TRO.2022.3168731}
}

@InProceedings{su2016learning,
    author="Su, Zhe
    and Kroemer, Oliver
    and Loeb, Gerald E.
    and Sukhatme, Gaurav S.
    and Schaal, Stefan",
    editor="Tuci, Elio
    and Giagkos, Alexandros
    and Wilson, Myra
    and Hallam, John",
    title="Learning to Switch Between Sensorimotor Primitives Using Multimodal Haptic Signals",
    booktitle="From Animals to Animats 14",
    year="2016",
    publisher="Springer International Publishing",
    address="Cham",
    pages="170--182",
    isbn="978-3-319-43488-9"
}

@ARTICLE{su2012compliance_biotac,
    AUTHOR={Su, Zhe  and Fishel, Jeremy A. and Yamamoto, Tomonori  and Loeb, Gerald E.},
    TITLE={Use of tactile feedback to control exploratory movements to characterize object compliance},
    JOURNAL={Frontiers in Neurorobotics},
    VOLUME={Volume 6 - 2012},
    YEAR={2012},
    URL={https://www.frontiersin.org/journals/neurorobotics/articles/10.3389/fnbot.2012.00007},
    DOI={10.3389/fnbot.2012.00007},
    ISSN={1662-5218},
}

@INPROCEEDINGS{yuan2017hardness,
  author={Yuan, Wenzhen and Zhu, Chenzhuo and Owens, Andrew and Srinivasan, Mandayam A. and Adelson, Edward H.},
  booktitle={2017 IEEE International Conference on Robotics and Automation (ICRA)}, 
  title={Shape-independent hardness estimation using deep learning and a GelSight tactile sensor}, 
  year={2017},
  volume={},
  number={},
  pages={951-958},
  doi={10.1109/ICRA.2017.7989116}
}

@INPROCEEDINGS{jin2025com,
  author={Jin, Shengmiao and Mo, Yuchen and Yuan, Wenzhen},
  booktitle={2025 IEEE International Conference on Robotics and Automation (ICRA)}, 
  title={Learning to Double Guess: An Active Perception Approach for Estimating the Center of Mass of Arbitrary Objects}, 
  year={2025},
  volume={},
  number={},
  pages={5843-5849},
  doi={10.1109/ICRA55743.2025.11127607}
}

@INPROCEEDINGS{Huang-RSS-22, 
    AUTHOR    = {Hung-Jui Huang AND Xiaofeng Guo AND Wenzhen Yuan}, 
    TITLE     = {{Understanding Dynamic Tactile Sensing for Liquid Property Estimation}}, 
    BOOKTITLE = {Proceedings of Robotics: Science and Systems}, 
    YEAR      = {2022}, 
    ADDRESS   = {New York City, NY, USA}, 
    MONTH     = {June}, 
    DOI       = {10.15607/RSS.2022.XVIII.072} 
}

@INPROCEEDINGS{guo2023estimating,
  author={Guo, Xiaofeng and Huang, Hung-Jui and Yuan, Wenzhen},
  booktitle={2023 IEEE/RSJ International Conference on Intelligent Robots and Systems (IROS)}, 
  title={Estimating Properties of Solid Particles Inside Container Using Touch Sensing}, 
  year={2023},
  volume={},
  number={},
  pages={8985-8992},
  doi={10.1109/IROS55552.2023.10341880}
}

@ARTICLE{seminara2019active,
    AUTHOR={Seminara, Lucia  and Gastaldo, Paolo  and Watt, Simon J.  and Valyear, Kenneth F.  and Zuher, Fernando  and Mastrogiovanni, Fulvio },
    TITLE={Active Haptic Perception in Robots: A Review},
    JOURNAL={Frontiers in Neurorobotics},
    VOLUME={Volume 13 - 2019},
    YEAR={2019},
    URL={https://www.frontiersin.org/journals/neurorobotics/articles/10.3389/fnbot.2019.00053},
    DOI={10.3389/fnbot.2019.00053},
    ISSN={1662-5218},
}

@INPROCEEDINGS{wang2020swingbot,
  author={Wang, Chen and Wang, Shaoxiong and Romero, Branden and Veiga, Filipe and Adelson, Edward},
  booktitle={2020 IEEE/RSJ International Conference on Intelligent Robots and Systems (IROS)}, 
  title={SwingBot: Learning Physical Features from In-hand Tactile Exploration for Dynamic Swing-up Manipulation}, 
  year={2020},
  volume={},
  number={},
  pages={5633-5640},
  doi={10.1109/IROS45743.2020.9341006}
}

@INPROCEEDINGS{huang2024kitchen,
  author={Huang, Hung-Jui and Xiang, Jingyi and Yuan, Wenzhen},
  booktitle={2024 IEEE International Conference on Robotics and Automation (ICRA)}, 
  title={Kitchen Artist: Precise Control of Liquid Dispensing for Gourmet Plating}, 
  year={2024},
  volume={},
  number={},
  pages={13933-13939},
  doi={10.1109/ICRA57147.2024.10611646}
}

@INPROCEEDINGS{luo2024pancake,
  author={Luo, Xinyuan and Jin, Shengmiao and Huang, Hung-Jui and Yuan, Wenzhen},
  booktitle={2024 IEEE/RSJ International Conference on Intelligent Robots and Systems (IROS)}, 
  title={An Intelligent Robotic System for Perceptive Pancake Batter Stirring and Precise Pouring}, 
  year={2024},
  volume={},
  number={},
  pages={5970-5977},
  doi={10.1109/IROS58592.2024.10802841}
}

@article{1-she2021cable,
    author = {Yu She and Shaoxiong Wang and Siyuan Dong and Neha Sunil and Alberto Rodriguez and Edward Adelson},
    title ={Cable manipulation with a tactile-reactive gripper},
    journal = {The International Journal of Robotics Research},
    volume = {40},
    number = {12-14},
    pages = {1385-1401},
    year = {2021},
    doi = {10.1177/02783649211027233},
}

@INPROCEEDINGS{2-dong2017improved,
  author={Dong, Siyuan and Yuan, Wenzhen and Adelson, Edward H.},
  booktitle={2017 IEEE/RSJ International Conference on Intelligent Robots and Systems (IROS)}, 
  title={Improved GelSight tactile sensor for measuring geometry and slip}, 
  year={2017},
  volume={},
  number={},
  pages={137-144},
  doi={10.1109/IROS.2017.8202149}
}

@ARTICLE{3-zhao2024tac,
  author={Zhao, Zihang and Li, Yuyang and Li, Wanlin and Qi, Zhenghao and Ruan, Lecheng and Zhu, Yixin and Althoefer, Kaspar},
  journal={IEEE Transactions on Robotics}, 
  title={Tac-Man: Tactile-Informed Prior-Free Manipulation of Articulated Objects}, 
  year={2025},
  volume={41},
  number={},
  pages={538-557},
  doi={10.1109/TRO.2024.3508134}
}

@INPROCEEDINGS{4-kim2022active,
  author={Kim, Sangwoon and Rodriguez, Alberto},
  booktitle={2022 International Conference on Robotics and Automation (ICRA)}, 
  title={Active Extrinsic Contact Sensing: Application to General Peg-in-Hole Insertion}, 
  year={2022},
  volume={},
  number={},
  pages={10241-10247},
  doi={10.1109/ICRA46639.2022.9812017}
}

@INPROCEEDINGS{5-hogan2020tactile,
  author={Hogan, Francois R. and Ballester, Jose and Dong, Siyuan and Rodriguez, Alberto},
  booktitle={2020 IEEE International Conference on Robotics and Automation (ICRA)}, 
  title={Tactile Dexterity: Manipulation Primitives with Tactile Feedback}, 
  year={2020},
  volume={},
  number={},
  pages={8863-8869},
  doi={10.1109/ICRA40945.2020.9196976}
}

@INPROCEEDINGS{9-wilson2023cable,
  author={Wilson, Achu and Jiang, Helen and Lian, Wenzhao and Yuan, Wenzhen},
  booktitle={2023 IEEE International Conference on Robotics and Automation (ICRA)}, 
  title={Cable Routing and Assembly using Tactile-driven Motion Primitives}, 
  year={2023},
  volume={},
  number={},
  pages={10408-10414},
  doi={10.1109/ICRA48891.2023.10161069}
}

@InProceedings{12-sunil2023visuotactile,
  title = 	 {Visuotactile Affordances for Cloth Manipulation with Local Control},
  author =       {Sunil, Neha and Wang, Shaoxiong and She, Yu and Adelson, Edward and Garcia, Alberto Rodriguez},
  booktitle = 	 {Proceedings of The 6th Conference on Robot Learning},
  pages = 	 {1596--1606},
  year = 	 {2023},
  editor = 	 {Liu, Karen and Kulic, Dana and Ichnowski, Jeff},
  volume = 	 {205},
  series = 	 {Proceedings of Machine Learning Research},
  month = 	 {14--18 Dec},
  publisher =    {PMLR},
  url = 	 {https://proceedings.mlr.press/v205/sunil23a.html},
}

@InProceedings{19-zhang2020towards,
  title = 	 {Towards Learning to Detect and Predict Contact Events on Vision-based Tactile Sensors},
  author =       {Zhang, Yazhan and Yuan, Weihao and Kan, Zicheng and Wang, Michael Yu},
  booktitle = 	 {Proceedings of the Conference on Robot Learning},
  pages = 	 {1395--1404},
  year = 	 {2020},
  editor = 	 {Kaelbling, Leslie Pack and Kragic, Danica and Sugiura, Komei},
  volume = 	 {100},
  series = 	 {Proceedings of Machine Learning Research},
  month = 	 {30 Oct--01 Nov},
  publisher =    {PMLR},
  url = 	 {https://proceedings.mlr.press/v100/zhang20b.html},
}

@INPROCEEDINGS{21-bronars2024texterity,
  author={Bronars, Antonia and Kim, Sangwoon and Patre, Parag and Rodriguez, Alberto},
  booktitle={2024 IEEE International Conference on Robotics and Automation (ICRA)}, 
  title={TEXterity: Tactile Extrinsic deXterity}, 
  year={2024},
  volume={},
  number={},
  pages={7976-7983},
  doi={10.1109/ICRA57147.2024.10610622}
}

@INPROCEEDINGS{27-xue2025reactive, 
    AUTHOR    = {Han Xue AND Jieji Ren AND Wendi Chen AND Gu Zhang AND Fang Yuan AND Guoying Gu AND Huazhe Xu AND Cewu Lu}, 
    TITLE     = {{Reactive Diffusion Policy: Slow-Fast Visual-Tactile Policy Learning for Contact-Rich Manipulation}}, 
    BOOKTITLE = {Proceedings of Robotics: Science and Systems}, 
    YEAR      = {2025}, 
    ADDRESS   = {LosAngeles, CA, USA}, 
    MONTH     = {June}, 
    DOI       = {10.15607/RSS.2025.XXI.052} 
}

@ARTICLE{28-chen2025dexforce,
  author={Chen, Claire and Yu, Zhongchun and Choi, Hojung and Cutkosky, Mark and Bohg, Jeannette},
  journal={IEEE Robotics and Automation Letters}, 
  title={DexForce: Extracting Force-Informed Actions From Kinesthetic Demonstrations for Dexterous Manipulation}, 
  year={2025},
  volume={10},
  number={6},
  pages={6416-6423},
  doi={10.1109/LRA.2025.3568318}
}

@ARTICLE{33-akinola2025tacsl,
  author={Akinola, Iretiayo and Xu, Jie and Carius, Jan and Fox, Dieter and Narang, Yashraj},
  journal={IEEE Transactions on Robotics}, 
  title={TacSL: A Library for Visuotactile Sensor Simulation and Learning}, 
  year={2025},
  volume={41},
  number={},
  pages={2645-2661},
  doi={10.1109/TRO.2025.3547267}
}

@InProceedings{34-huang20253d,
  title = 	 {3D-ViTac: Learning Fine-Grained Manipulation with Visuo-Tactile Sensing},
  author =       {Huang, Binghao and Wang, Yixuan and Yang, Xinyi and Luo, Yiyue and Li, Yunzhu},
  booktitle = 	 {Proceedings of The 8th Conference on Robot Learning},
  pages = 	 {2557--2578},
  year = 	 {2025},
  editor = 	 {Agrawal, Pulkit and Kroemer, Oliver and Burgard, Wolfram},
  volume = 	 {270},
  series = 	 {Proceedings of Machine Learning Research},
  month = 	 {06--09 Nov},
  publisher =    {PMLR},
  url = 	 {https://proceedings.mlr.press/v270/huang25e.html},
}

@InProceedings{38-guzey2023dexterity,
  title = 	 {Dexterity from Touch: Self-Supervised Pre-Training of Tactile Representations with Robotic Play},
  author =       {Guzey, Irmak and Evans, Ben and Chintala, Soumith and Pinto, Lerrel},
  booktitle = 	 {Proceedings of The 7th Conference on Robot Learning},
  pages = 	 {3142--3166},
  year = 	 {2023},
  editor = 	 {Tan, Jie and Toussaint, Marc and Darvish, Kourosh},
  volume = 	 {229},
  series = 	 {Proceedings of Machine Learning Research},
  month = 	 {06--09 Nov},
  publisher =    {PMLR},
  url = 	 {https://proceedings.mlr.press/v229/guzey23a.html},
}

@INPROCEEDINGS{41-yu2024octopi, 
    AUTHOR    = {Samson Yu AND Lin Kelvin AND Anxing Xiao AND Jiafei Duan AND Harold Soh}, 
    TITLE     = {{Octopi: Object Property Reasoning with Large Tactile-Language Models}}, 
    BOOKTITLE = {Proceedings of Robotics: Science and Systems}, 
    YEAR      = {2024}, 
    ADDRESS   = {Delft, Netherlands}, 
    MONTH     = {July}, 
    DOI       = {10.15607/RSS.2024.XX.066} 
}

@inproceedings{43-fenganytouch,
 author = {Feng, Ruoxuan and Hu, Jiangyu and Xia, Wenke and Gao, Tianci and Shen, Ao and Sun, Yuhao and Fang, Bin and Hu, Di},
 booktitle = {International Conference on Learning Representations},
 editor = {Y. Yue and A. Garg and N. Peng and F. Sha and R. Yu},
 pages = {31265--31285},
 title = {AnyTouch: Learning Unified Static-Dynamic Representation across Multiple Visuo-tactile Sensors},
 url = {https://proceedings.iclr.cc/paper_files/paper/2025/file/4d893f766ab60e5337659b9e71883af4-Paper-Conference.pdf},
 volume = {2025},
 year = {2025}
}

@InProceedings{46-zhao2025transferable,
  title = 	 {Transferable Tactile Transformers for Representation Learning Across Diverse Sensors and Tasks},
  author =       {Zhao, Jialiang and Ma, Yuxiang and Wang, Lirui and Adelson, Edward},
  booktitle = 	 {Proceedings of The 8th Conference on Robot Learning},
  pages = 	 {3766--3779},
  year = 	 {2025},
  editor = 	 {Agrawal, Pulkit and Kroemer, Oliver and Burgard, Wolfram},
  volume = 	 {270},
  series = 	 {Proceedings of Machine Learning Research},
  month = 	 {06--09 Nov},
  publisher =    {PMLR},
  url = 	 {https://proceedings.mlr.press/v270/zhao25c.html},
}

@INPROCEEDINGS{47-romero2024eyesight,
  author={Romero, Branden and Fang, Hao-Shu and Agrawal, Pulkit and Adelson, Edward},
  booktitle={2024 IEEE/RSJ International Conference on Intelligent Robots and Systems (IROS)}, 
  title={EyeSight Hand: Design of a Fully-Actuated Dexterous Robot Hand with Integrated Vision-Based Tactile Sensors and Compliant Actuation}, 
  year={2024},
  volume={},
  number={},
  pages={1853-1860},
  doi={10.1109/IROS58592.2024.10802778}
}

@inproceedings{56-zhu2025touch,
 author = {Zhu, Xinyue and Huang, Binghao and Li, Yunzhu},
 booktitle = {Advances in Neural Information Processing Systems},
 doi = {10.52202/085713-5143},
 editor = {D. Belgrave and C. Zhang and H. Lin and R. Pascanu and P. Koniusz and M. Ghassemi and N. Chen},
 pages = {153783--153812},
 publisher = {Curran Associates, Inc.},
 title = {Touch in the Wild: Learning Fine-Grained Manipulation with a Portable Visuo-Tactile Gripper},
 url = {https://proceedings.neurips.cc/paper_files/paper/2025/file/e1fcd183ab33714a8464e4e9a20ac710-Paper-Conference.pdf},
 volume = {38, Main Conference},
 year = {2025}
}

@INPROCEEDINGS{64-wu2025canonical,
  author={Wu, Tianhao and Li, Jinzhou and Zhang, Jiyao and Wu, Mingdong and Dong, Hao},
  booktitle={2025 IEEE International Conference on Robotics and Automation (ICRA)}, 
  title={Canonical Representation and Force-Based Pretraining of 3D Tactile for Dexterous Visuo-Tactile Policy Learning}, 
  year={2025},
  volume={},
  number={},
  pages={6786-6792},
  doi={10.1109/ICRA55743.2025.11128094}
}

@InProceedings{65-xu2025exumi,
  title = 	 {exUMI: Extensible Robot Teaching System with Action-aware Task-agnostic Tactile Representation},
  author =       {Xu, Yue and Wei, Litao and An, Pengyu and Zhang, Qingyu and Li, Yong-Lu},
  booktitle = 	 {Proceedings of The 9th Conference on Robot Learning},
  pages = 	 {2536--2554},
  year = 	 {2025},
  editor = 	 {Lim, Joseph and Song, Shuran and Park, Hae-Won},
  volume = 	 {305},
  series = 	 {Proceedings of Machine Learning Research},
  month = 	 {27--30 Sep},
  publisher =    {PMLR},
  url = 	 {https://proceedings.mlr.press/v305/xu25e.html},
}

@INPROCEEDINGS{67-ACP,
  author={Hou, Yifan and Liu, Zeyi and Chi, Cheng and Cousineau, Eric and Kuppuswamy, Naveen and Feng, Siyuan and Burchfiel, Benjamin and Song, Shuran},
  booktitle={2025 IEEE International Conference on Robotics and Automation (ICRA)}, 
  title={Adaptive Compliance Policy: Learning Approximate Compliance for Diffusion Guided Control}, 
  year={2025},
  volume={},
  number={},
  pages={4829-4836},
  doi={10.1109/ICRA55743.2025.11128452}
}

@INPROCEEDINGS{68-liu2025factr, 
    AUTHOR    = {Jason Jingzhou Liu AND Yulong Li AND Kenneth Shaw AND Tony Tao AND Russ Salakhutdinov AND Deepak Pathak}, 
    TITLE     = {{FACTR: Force-Attending Curriculum Training for Contact-Rich Policy Learning}}, 
    BOOKTITLE = {Proceedings of Robotics: Science and Systems}, 
    YEAR      = {2025}, 
    ADDRESS   = {LosAngeles, CA, USA}, 
    MONTH     = {June}, 
    DOI       = {10.15607/RSS.2025.XXI.079} 
}

@INPROCEEDINGS{69-jones2025beyond,
  author={Jones, Joshua and Mees, Oier and Sferrazza, Carmelo and Stachowicz, Kyle and Abbeel, Pieter and Levine, Sergey},
  booktitle={2025 IEEE International Conference on Robotics and Automation (ICRA)}, 
  title={Beyond Sight: Finetuning Generalist Robot Policies with Heterogeneous Sensors via Language Grounding}, 
  year={2025},
  volume={},
  number={},
  pages={5961-5968},
  doi={10.1109/ICRA55743.2025.11127987}
}

@INPROCEEDINGS{71-ForceMimic,
  author={Liu, Wenhai and Wang, Junbo and Wang, Yiming and Wang, Weiming and Lu, Cewu},
  booktitle={2025 IEEE International Conference on Robotics and Automation (ICRA)}, 
  title={ForceMimic: Force-Centric Imitation Learning with Force-Motion Capture System for Contact-Rich Manipulation}, 
  year={2025},
  volume={},
  number={},
  pages={1105-1112},
  doi={10.1109/ICRA55743.2025.11128061}
}

@INPROCEEDINGS{78-george2025vital,
  author={George, Abraham and Gano, Selam and Katragadda, Pranav and Farimani, Amir Barati},
  booktitle={2025 IEEE International Conference on Robotics and Automation (ICRA)}, 
  title={VITaL Pretraining: Visuo-Tactile Pretraining for Tactile and Non-Tactile Manipulation Policies}, 
  year={2025},
  volume={},
  number={},
  pages={258-264},
  doi={10.1109/ICRA55743.2025.11128336}
}

@ARTICLE{80-he2025foar,
  author={He, Zihao and Fang, Hongjie and Chen, Jingjing and Fang, Hao-Shu and Lu, Cewu},
  journal={IEEE Robotics and Automation Letters}, 
  title={FoAR: Force-Aware Reactive Policy for Contact-Rich Robotic Manipulation}, 
  year={2025},
  volume={10},
  number={6},
  pages={5625-5632},
  doi={10.1109/LRA.2025.3560871}
}

@inproceedings{86-wu2025freetacman,
  title={FreeTacMan: Robot-free Visuo-Tactile Data Collection System for Contact-rich Manipulation},
  author={Wu, Longyan and Yu, Checheng and Ren, Jieji and Chen, Li and Jiang, Yufei and Huang, Ran and Gu, Guoying and Li, Hongyang},
  booktitle={2026 IEEE International Conference on Robotics and Automation (ICRA)},
  year={2026}
}

@InProceedings{88-higuera2025sparsh,
  title = 	 {Sparsh: Self-supervised touch representations for vision-based tactile sensing},
  author =       {Higuera, Carolina and Sharma, Akash and Bodduluri, Chaithanya Krishna and Fan, Taosha and Lancaster, Patrick and Kalakrishnan, Mrinal and Kaess, Michael and Boots, Byron and Lambeta, Mike and Wu, Tingfan and Mukadam, Mustafa},
  booktitle = 	 {Proceedings of The 8th Conference on Robot Learning},
  pages = 	 {885--915},
  year = 	 {2025},
  editor = 	 {Agrawal, Pulkit and Kroemer, Oliver and Burgard, Wolfram},
  volume = 	 {270},
  series = 	 {Proceedings of Machine Learning Research},
  month = 	 {06--09 Nov},
  publisher =    {PMLR},
  url = 	 {https://proceedings.mlr.press/v270/higuera25a.html},
}

@inproceedings{89-zhang2026touchguide,
  title={TouchGuide: Inference-Time Steering of Visuomotor Policies via Touch Guidance},
  author={Zhang, Zhemeng and Ma, Jiahua and Yang, Xincheng and Wen, Xin and Zhang, Yuzhi and Li, Boyan and Qin, Yiran and Liu, Jin and Zhao, Can and Kang, Li and others},
  booktitle={Proceedings of Robotics: Science and Systems},
  year={2026}
}

@inproceedings{90-chen2026multimodal,
  title={Multi-Modal Manipulation via Multi-Modal Policy Consensus},
  author={Chen, Haonan and Xu, Jiaming and Chen, Hongyu and Hong, Kaiwen and Huang, Binghao and Liu, Chaoqi and Mao, Jiayuan and Li, Yunzhu and Du, Yilun and Driggs-Campbell, Katherine},
  booktitle={2026 IEEE International Conference on Robotics and Automation (ICRA)},
  year={2026}
}

@ARTICLE{91-xu2025unit,
  author={Xu, Zhengtong and Uppuluri, Raghava and Zhang, Xinwei and Fitch, Cael and Crandall, Philip Glen and Shou, Wan and Wang, Dongyi and She, Yu},
  journal={IEEE Robotics and Automation Letters}, 
  title={UniT: Data Efficient Tactile Representation With Generalization to Unseen Objects}, 
  year={2025},
  volume={10},
  number={6},
  pages={5481-5488},
  doi={10.1109/LRA.2025.3559835}
}

@inproceedings{92-sitr,
 author = {Gupta, Harsh and Mo, Yuchen and Jin, Shengmiao and Yuan, Wenzhen},
 booktitle = {International Conference on Learning Representations},
 editor = {Y. Yue and A. Garg and N. Peng and F. Sha and R. Yu},
 pages = {96515--96538},
 title = {Sensor-Invariant Tactile Representation},
 url = {https://proceedings.iclr.cc/paper_files/paper/2025/file/f05a11e682bf9c57ab62f7cff08fbe5b-Paper-Conference.pdf},
 volume = {2025},
 year = {2025}
}

@article{huang2026flexitac,
  title={{FlexiTac: A Low-Cost, Open-Source, Scalable Tactile Sensing Solution for Robotic Systems}},
  author={Huang, Binghao and Li, Yunzhu},
  journal={arXiv preprint arXiv:2604.28156},
  year={2026}
}

@inproceedings{98-FARM,
  title={Tactile-Conditioned Diffusion Policy for Force-Aware Robotic Manipulation},
  author={Helmut, Erik and Funk, Niklas and Schneider, Tim and de Farias, Cristiana and Peters, Jan},
  booktitle={2026 IEEE International Conference on Robotics and Automation (ICRA)},
  year={2026}
}

@ARTICLE{99-zhou2025admittance,
  author={Zhou, Bo and Jiao, Ruixuan and Li, Yi and Yuan, Xiaogang and Fang, Fang and Li, Shihua},
  journal={IEEE Transactions on Industrial Electronics}, 
  title={Admittance Visuomotor Policy Learning for General-Purpose Contact-Rich Manipulations}, 
  year={2025},
  volume={72},
  number={11},
  pages={11520-11529},
  doi={10.1109/TIE.2025.3561816}
}

@inproceedings{100-CR-DAgger,
 author = {Xu, Xiaomeng and Hou, Yifan and Liu, Zeyi and Song, Shuran},
 booktitle = {Advances in Neural Information Processing Systems},
 doi = {10.52202/085713-4660},
 editor = {D. Belgrave and C. Zhang and H. Lin and R. Pascanu and P. Koniusz and M. Ghassemi and N. Chen},
 pages = {139559--139581},
 publisher = {Curran Associates, Inc.},
 title = {Compliant Residual DAgger: Improving Real-World Contact-Rich Manipulation with Human Corrections},
 url = {https://proceedings.neurips.cc/paper_files/paper/2025/file/cbf883a744952d4a40591271a58ab9d0-Paper-Conference.pdf},
 volume = {38, Main Conference},
 year = {2025}
}

@inproceedings{101-UMI-FT,
  title={In-the-Wild Compliant Manipulation with UMI-FT},
  author={Choi, Hojung and Hou, Yifan and Pan, Chuer and Hong, Seongheon and Patel, Austin and Xu, Xiaomeng and Cutkosky, Mark R and Song, Shuran},
  booktitle={2026 IEEE International Conference on Robotics and Automation (ICRA)},
  year={2026}
}

@InProceedings{102-Sparsh-X,
  title = 	 {Tactile Beyond Pixels: Multisensory Touch Representations for Robot Manipulation},
  author =       {Higuera, Carolina and Sharma, Akash and Fan, Taosha and Bodduluri, Chaithanya Krishna and Boots, Byron and Kaess, Michael and Lambeta, Mike and Wu, Tingfan and Liu, Zixi and Hogan, Francois Robert and Mukadam, Mustafa},
  booktitle = 	 {Proceedings of The 9th Conference on Robot Learning},
  pages = 	 {105--123},
  year = 	 {2025},
  editor = 	 {Lim, Joseph and Song, Shuran and Park, Hae-Won},
  volume = 	 {305},
  series = 	 {Proceedings of Machine Learning Research},
  month = 	 {27--30 Sep},
  publisher =    {PMLR},
  url = 	 {https://proceedings.mlr.press/v305/higuera25a.html},
}

@INPROCEEDINGS{Comp-ACT,
  author={Kamijo, Tatsuya and Beltran-Hernandez, Cristian C. and Hamaya, Masashi},
  booktitle={2024 IEEE/RSJ International Conference on Intelligent Robots and Systems (IROS)}, 
  title={Learning Variable Compliance Control From a Few Demonstrations for Bimanual Robot with Haptic Feedback Teleoperation System}, 
  year={2024},
  volume={},
  number={},
  pages={12663-12670},
  doi={10.1109/IROS58592.2024.10801731}
}

@article{Tactile-VLA,
  title={{Tactile-VLA: Unlocking Vision-Language-Action Model's Physical Knowledge for Tactile Generalization}},
  author={Huang, Jialei and Wang, Shuo and Lin, Fanqi and Hu, Yihang and Wen, Chuan and Gao, Yang},
  journal={arXiv preprint arXiv:2507.09160},
  year={2025}
}

@ARTICLE{VLA-Touch,
  author={Bi, Jianxin and Ma, Kevin Yuchen and Hao, Ce and Zheng, Mike Shou and Soh, Harold},
  journal={IEEE Robotics and Automation Letters}, 
  title={VLA-Touch: Enhancing Vision-Language-Action Model With Dual-Level Tactile Feedback}, 
  year={2026},
  volume={11},
  number={7},
  pages={8487-8494},
  doi={10.1109/LRA.2026.3692345}
}

@inproceedings{MLA,
  title={MLA: A Multisensory Language-Action Model for Multimodal Understanding and Forecasting in Robotic Manipulation},
  author={Liu, Zhuoyang and Liu, Jiaming and Xu, Jiadong and Han, Nuowei and Gu, Chenyang and Chen, Hao and Zhou, Kaichen and Zhang, Renrui and Hsieh, Kai Chin and Wu, Kun and others},
  booktitle={2026 IEEE International Conference on Robotics and Automation (ICRA)},
  year={2026}
}

@inproceedings{TacVLA,
    title={TacVLA: Contact-Aware Tactile Fusion for Robust Vision-Language-Action Manipulation},
    author={Zhang, Kaidi and Zhang, Heng and Xu, Zhengtong and Zhang, Zhiyuan and Prince, Md Rakibul Islam and Li, Xiang and Han, Xiaojing and Zhou, Yuhao and Ajoudani, Arash and She, Yu},
    booktitle={2026 IEEE/RSJ International Conference on Intelligent Robots and Systems (IROS)},
    year={2026}
}

@inproceedings{ForceVLA,
 author = {Yu, Jiawen and Liu, Hairuo and Yu, Qiaojun and Ren, Jieji and Hao, Ce and Ding, Haitong and Huang, Guangyu and Huang, Guofan and Song, Yan and Cai, Panpan and Zhang, Wenqiang and Lu, Cewu},
 booktitle = {Advances in Neural Information Processing Systems},
 doi = {10.52202/085713-3124},
 editor = {D. Belgrave and C. Zhang and H. Lin and R. Pascanu and P. Koniusz and M. Ghassemi and N. Chen},
 pages = {93409--93439},
 publisher = {Curran Associates, Inc.},
 title = {ForceVLA: Enhancing VLA Models with a Force-aware MoE for Contact-rich Manipulation},
 url = {https://proceedings.neurips.cc/paper_files/paper/2025/file/8633b46e12cc5f2ee1f05a6ca2c65b38-Paper-Conference.pdf},
 volume = {38, Main Conference},
 year = {2025}
}

@InProceedings{TA-VLA,
  title = 	 {Elucidating the Design Space of Torque-aware Vision-Language-Action Models},
  author =       {Zhang, Zongzheng and Xu, Haobo and Yang, Zhuo and Yue, Chenghao and Lin, Zehao and Gao, Huan-ang and Wang, Ziwei and Zhao, Hao},
  booktitle = 	 {Proceedings of The 9th Conference on Robot Learning},
  pages = 	 {4019--4037},
  year = 	 {2025},
  editor = 	 {Lim, Joseph and Song, Shuran and Park, Hae-Won},
  volume = 	 {305},
  series = 	 {Proceedings of Machine Learning Research},
  month = 	 {27--30 Sep},
  publisher =    {PMLR},
  url = 	 {https://proceedings.mlr.press/v305/zhang25k.html},
}

@inproceedings{FD-VLA,
  title={FD-VLA: Force-Distilled Vision-Language-Action Model for Contact-Rich Manipulation},
  author={Zhao, Ruiteng and Wang, Wenshuo and Ma, Yicheng and Li, Xiaocong and Tay, Francis EH and Ang Jr, Marcelo H and Zhu, Haiyue},
  booktitle={2026 IEEE International Conference on Robotics and Automation (ICRA)},
  year={2026}
}

@InProceedings{ForceVLA2,
    author    = {Li, Yang and Zhaxizhuoma, Zhaxizhuoma and Jiang, Hongru and Xia, Junjie and Zhang, Hongquan and Du, Jinda and Zhou, Yunsong and Zeng, Jia and Hao, Ce and Ren, Jieji and Yu, Qiaojun and Lu, Cewu and Qiao, Yu and Pang, Jiangmiao},
    title     = {ForceVLA2: Unleashing Hybrid Force-Position Control with Force Awareness for Contact-Rich Manipulation},
    booktitle = {Proceedings of the IEEE/CVF Conference on Computer Vision and Pattern Recognition (CVPR)},
    month     = {June},
    year      = {2026},
    pages     = {8911-8920}
}

@InProceedings{AT-VLA,
    author    = {Li, Xiaoqi and Cai, Muhe and Xu, Jiadong and Zhu, Juan and Fan, Hongwei and Shen, Yan and Ren, Guangrui and Dong, Hao},
    title     = {AT-VLA: Adaptive Tactile Injection for Enhanced Feedback Reaction in Vision-Language-Action Models},
    booktitle = {Proceedings of the IEEE/CVF Conference on Computer Vision and Pattern Recognition (CVPR)},
    month     = {June},
    year      = {2026},
    pages     = {28764-28774}
}

@article{T-Rex,
  title={{T-Rex: Tactile-Reactive Dexterous Manipulation}},
  author={Niu, Dantong and Liu, Zhuoyang and Wang, Zekai and Shao, Boning and Yin, Zhao-Heng and Pai, Anirudh and Sharma, Yuvan and Saravalle, Stefano and Zheng, Ruijie and Wang, Jing and others},
  journal={arXiv preprint arXiv:2606.17055},
  year={2026}
}

@article{FTP-1,
  title={{FTP-1: A Generalist Foundation Tactile Policy Across Tactile Sensors for Contact-Rich Manipulation}},
  author={Yuan, Chengbo and Zhang, Zicheng and Zhou, Mingjie and Chen, Wendi and Wang, Yi and Liu, Zhuoyang and Niu, Dantong and Wang, Shuo and Zhang, Hui and Zhang, Wenkang and others},
  journal={arXiv preprint arXiv:2606.13102},
  year={2026}
}

@inproceedings{guo2025ctrl,
 author = {Guo, Yanjiang and Shi, Lucy and Chen, Jianyu and Finn, Chelsea},
 booktitle = {International Conference on Learning Representations},
 editor = {C. Vondrick and B. Hariharan and C. Raffel and L. Pinto and D. Yang and A. Faust},
 pages = {6121--6138},
 title = {Ctrl-World: A Controllable Generative World Model for Robot Manipulation},
 url = {https://proceedings.iclr.cc/paper_files/paper/2026/file/0ae94013da7cd459402fd77874e09ee3-Paper-Conference.pdf},
 volume = {2026},
 year = {2026}
}

@InProceedings{wu2023daydreamer,
  title = 	 {DayDreamer: World Models for Physical Robot Learning},
  author =       {Wu, Philipp and Escontrela, Alejandro and Hafner, Danijar and Abbeel, Pieter and Goldberg, Ken},
  booktitle = 	 {Proceedings of The 6th Conference on Robot Learning},
  pages = 	 {2226--2240},
  year = 	 {2023},
  editor = 	 {Liu, Karen and Kulic, Dana and Ichnowski, Jeff},
  volume = 	 {205},
  series = 	 {Proceedings of Machine Learning Research},
  month = 	 {14--18 Dec},
  publisher =    {PMLR},
  url = 	 {https://proceedings.mlr.press/v205/wu23c.html},
}

@article{higuera2026visuotactileworldmodels,
  title={{Visuo-Tactile World Models}},
  author={Higuera, Carolina and Arnaud, Sergio and Boots, Byron and Mukadam, Mustafa and Hogan, Francois Robert and Meier, Franziska},
  journal={arXiv preprint arXiv:2602.06001},
  year={2026}
}

@article{zheng2026omnivtavisuotactileworldmodeling,
  title={{OmniVTA: Visuo-Tactile World Modeling for Contact-Rich Robotic Manipulation}},
  author={Zheng, Yuhang and Gu, Songen and Li, Weize and Zheng, Yupeng and Zang, Yujie and Tian, Shuai and Li, Xiang and Hao, Ce and Gao, Chen and Liu, Si and others},
  journal={arXiv preprint arXiv:2603.19201},
  year={2026}
}

@article{ye2026learningfeelfuturedreamtacvla,
  title={{Learning to Feel the Future: DreamTacVLA for Contact-Rich Manipulation}},
  author={Ye, Guo and Zhang, Zexi and Zhao, Xu and Wu, Shang and Lu, Haoran and Lu, Shihan and Liu, Han},
  journal={arXiv preprint arXiv:2512.23864},
  year={2026}
}

@article{niu2026learningversatilehumanoidmanipulation,
  title={{Learning Versatile Humanoid Manipulation with Touch Dreaming}},
  author={Niu, Yaru and Fang, Zhenlong and Chen, Binghong and Zhou, Shuai and Senthilkumaran, Revanth Krishna and Zhang, Hao and Chen, Bingqing and Qiu, Chen and Tseng, H Eric and Francis, Jonathan and others},
  journal={arXiv preprint arXiv:2604.13015},
  year={2026}
}

@inproceedings{heng2025vitacformer,
  title={ViTacFormer: Learning Cross-Modal Representation for Visuo-Tactile Dexterous Manipulation},
  author={Heng, Liang and Geng, Haoran and Zhang, Kaifeng and Abbeel, Pieter and Malik, Jitendra},
  booktitle={Proceedings of Robotics: Science and Systems},
  year={2026}
}

@INPROCEEDINGS{tian2019manipulationbyfeel,
  author={Tian, Stephen and Ebert, Frederik and Jayaraman, Dinesh and Mudigonda, Mayur and Finn, Chelsea and Calandra, Roberto and Levine, Sergey},
  booktitle={2019 International Conference on Robotics and Automation (ICRA)}, 
  title={Manipulation by Feel: Touch-Based Control with Deep Predictive Models}, 
  year={2019},
  volume={},
  number={},
  pages={818-824},
  doi={10.1109/ICRA.2019.8794219}
}

@INPROCEEDINGS{Sutanto_2019,
  author={Sutanto, Giovanni and Ratliff, Nathan and Sundaralingam, Balakumar and Chebotar, Yevgen and Su, Zhe and Handa, Ankur and Fox, Dieter},
  booktitle={2019 International Conference on Robotics and Automation (ICRA)}, 
  title={Learning Latent Space Dynamics for Tactile Servoing}, 
  year={2019},
  volume={},
  number={},
  pages={3622-3628},
  doi={10.1109/ICRA.2019.8793520}
}

@ARTICLE{xu2024letacmpc,
  author={Xu, Zhengtong and She, Yu},
  journal={IEEE Transactions on Robotics}, 
  title={LeTac-MPC: Learning Model Predictive Control for Tactile-Reactive Grasping}, 
  year={2024},
  volume={40},
  number={},
  pages={4376-4395},
  doi={10.1109/TRO.2024.3463470}
}

@INPROCEEDINGS{ai2024robopack, 
    AUTHOR    = {Bo Ai AND Stephen Tian AND Haochen Shi AND Yixuan Wang AND Cheston Tan AND Yunzhu Li AND Jiajun Wu}, 
    TITLE     = {{RoboPack: Learning Tactile-Informed Dynamics Models for Dense Packing}}, 
    BOOKTITLE = {Proceedings of Robotics: Science and Systems}, 
    YEAR      = {2024}, 
    ADDRESS   = {Delft, Netherlands}, 
    MONTH     = {July}, 
    DOI       = {10.15607/RSS.2024.XX.130} 
}

@article{richalet1978mph,
  title={Model predictive heuristic control: Applications to industrial processes},
  author={Richalet, Jacques and Rault, A and Testud, JL and Papon, J},
  journal={Automatica},
  volume={14},
  number={5},
  pages={413--428},
  year={1978},
  publisher={Elsevier}
}

@article{mayne2000constrained,
    title = {Constrained model predictive control: Stability and optimality},
    journal = {Automatica},
    volume = {36},
    number = {6},
    pages = {789-814},
    year = {2000},
    issn = {0005-1098},
    doi = {https://doi.org/10.1016/S0005-1098(99)00214-9},
    url = {https://www.sciencedirect.com/science/article/pii/S0005109899002149},
    author = {D.Q. Mayne and J.B. Rawlings and C.V. Rao and P.O.M. Scokaert},
}

@article{dream-tac,
  title={{Dream-Tac: A Unified Tactile World Action Model for Contact-Rich Robot Manipulation}},
  author={Lou, Yunfan and Ye, Yifan and Fu, Yankai and Cen, Jun and Chi, Xiaowei and Lyu, Yaoxu and Jia, Peidong and Han, Sirui and Lu, Zhihe and Zhang, Shanghang},
  journal={arXiv preprint arXiv:2606.08737},
  year={2026}
}

@INPROCEEDINGS{khandate2023sampling, 
    AUTHOR    = {Gagan Khandate AND Siqi Shang AND Eric T Chang AND Tristan L Saidi AND Johnson Adams AND Matei Ciocarlie}, 
    TITLE     = {{Sampling-based Exploration for Reinforcement Learning of Dexterous Manipulation}}, 
    BOOKTITLE = {Proceedings of Robotics: Science and Systems}, 
    YEAR      = {2023}, 
    ADDRESS   = {Daegu, Republic of Korea}, 
    MONTH     = {July}, 
    DOI       = {10.15607/RSS.2023.XIX.020} 
}

@inproceedings{miller2026enhancing,
 author = {Miller, Elle and McInroe, Trevor and Abel, David and Mac Aodha, Oisin and Vijayakumar, Sethu},
 booktitle = {Advances in Neural Information Processing Systems},
 doi = {10.52202/085713-4312},
 editor = {D. Belgrave and C. Zhang and H. Lin and R. Pascanu and P. Koniusz and M. Ghassemi and N. Chen},
 pages = {129460--129494},
 publisher = {Curran Associates, Inc.},
 title = {Enhancing Tactile-based Reinforcement Learning for Robotic Control},
 url = {https://proceedings.neurips.cc/paper_files/paper/2025/file/bc09efb501c801ed92e181e26a885c2d-Paper-Conference.pdf},
 volume = {38, Main Conference},
 year = {2025}
}

@INPROCEEDINGS{van2015learning,
  author={van Hoof, Herke and Hermans, Tucker and Neumann, Gerhard and Peters, Jan},
  booktitle={2015 IEEE-RAS 15th International Conference on Humanoid Robots (Humanoids)}, 
  title={Learning robot in-hand manipulation with tactile features}, 
  year={2015},
  volume={},
  number={},
  pages={121-127},
  doi={10.1109/HUMANOIDS.2015.7363524}
}

@INPROCEEDINGS{chebotar2016self,
  author={Chebotar, Yevgen and Hausman, Karol and Su, Zhe and Sukhatme, Gaurav S. and Schaal, Stefan},
  booktitle={2016 IEEE/RSJ International Conference on Intelligent Robots and Systems (IROS)}, 
  title={Self-supervised regrasping using spatio-temporal tactile features and reinforcement learning}, 
  year={2016},
  volume={},
  number={},
  pages={1960-1966},
  doi={10.1109/IROS.2016.7759309}
}

@ARTICLE{church2020deep,
  author={Church, Alex and Lloyd, John and Hadsell, Raia and Lepora, Nathan F.},
  journal={IEEE Robotics and Automation Letters}, 
  title={Deep Reinforcement Learning for Tactile Robotics: Learning to Type on a Braille Keyboard}, 
  year={2020},
  volume={5},
  number={4},
  pages={6145-6152},
  doi={10.1109/LRA.2020.3010461}
}

@InProceedings{30-qi2023general,
  title = 	 {General In-hand Object Rotation with Vision and Touch},
  author =       {Qi, Haozhi and Yi, Brent and Suresh, Sudharshan and Lambeta, Mike and Ma, Yi and Calandra, Roberto and Malik, Jitendra},
  booktitle = 	 {Proceedings of The 7th Conference on Robot Learning},
  pages = 	 {2549--2564},
  year = 	 {2023},
  editor = 	 {Tan, Jie and Toussaint, Marc and Darvish, Kourosh},
  volume = 	 {229},
  series = 	 {Proceedings of Machine Learning Research},
  month = 	 {06--09 Nov},
  publisher =    {PMLR},
  url = 	 {https://proceedings.mlr.press/v229/qi23a.html},
}

@ARTICLE{31-lee2020making,
  author={Lee, Michelle A. and Zhu, Yuke and Zachares, Peter and Tan, Matthew and Srinivasan, Krishnan and Savarese, Silvio and Fei-Fei, Li and Garg, Animesh and Bohg, Jeannette},
  journal={IEEE Transactions on Robotics}, 
  title={Making Sense of Vision and Touch: Learning Multimodal Representations for Contact-Rich Tasks}, 
  year={2020},
  volume={36},
  number={3},
  pages={582-596},
  doi={10.1109/TRO.2019.2959445}
}

@INPROCEEDINGS{32-yin2023rotating, 
    AUTHOR    = {Zhao-Heng Yin AND Binghao Huang AND Yuzhe Qin AND Qifeng Chen AND Xiaolong Wang}, 
    TITLE     = {{Rotating without Seeing: Towards In-hand Dexterity through Touch}}, 
    BOOKTITLE = {Proceedings of Robotics: Science and Systems}, 
    YEAR      = {2023}, 
    ADDRESS   = {Daegu, Republic of Korea}, 
    MONTH     = {July}, 
    DOI       = {10.15607/RSS.2023.XIX.036} 
}

@INPROCEEDINGS{35-yuan2024robot,
  author={Yuan, Ying and Che, Haichuan and Qin, Yuzhe and Huang, Binghao and Yin, Zhao-Heng and Lee, Kang-Won and Wu, Yi and Lim, Soo-Chul and Wang, Xiaolong},
  booktitle={2024 IEEE International Conference on Robotics and Automation (ICRA)}, 
  title={Robot Synesthesia: In-Hand Manipulation with Visuotactile Sensing}, 
  year={2024},
  volume={},
  number={},
  pages={6558-6565},
  doi={10.1109/ICRA57147.2024.10610532}
}

@INPROCEEDINGS{36-dong2021tactile,
  author={Dong, Siyuan and Jha, Devesh K. and Romeres, Diego and Kim, Sangwoon and Nikovski, Daniel and Rodriguez, Alberto},
  booktitle={2021 IEEE International Conference on Robotics and Automation (ICRA)}, 
  title={Tactile-RL for Insertion: Generalization to Objects of Unknown Geometry}, 
  year={2021},
  volume={},
  number={},
  pages={6437-6443},
  doi={10.1109/ICRA48506.2021.9561646}
}

@InProceedings{45-yang2025anyrotate,
  title = 	 {AnyRotate: Gravity-Invariant In-Hand Object Rotation with Sim-to-Real Touch},
  author =       {Yang, Max and lu, chenghua and Church, Alex and Lin, Yijiong and Ford, Christopher J. and Li, Haoran and Psomopoulou, Efi and Barton, David A.W. and Lepora, Nathan F.},
  booktitle = 	 {Proceedings of The 8th Conference on Robot Learning},
  pages = 	 {4727--4747},
  year = 	 {2025},
  editor = 	 {Agrawal, Pulkit and Kroemer, Oliver and Burgard, Wolfram},
  volume = 	 {270},
  series = 	 {Proceedings of Machine Learning Research},
  month = 	 {06--09 Nov},
  publisher =    {PMLR},
  url = 	 {https://proceedings.mlr.press/v270/yang25c.html},
}

@INPROCEEDINGS{49-pitz2023dextrous,
  author={Pitz, Johannes and Röstel, Lennart and Sievers, Leon and Bäuml, Berthold},
  booktitle={2023 IEEE International Conference on Robotics and Automation (ICRA)}, 
  title={Dextrous Tactile In-Hand Manipulation Using a Modular Reinforcement Learning Architecture}, 
  year={2023},
  volume={},
  number={},
  pages={1852-1858},
  doi={10.1109/ICRA48891.2023.10160756}
}

@INPROCEEDINGS{63-yin2025learning,
  author={Yin, Jessica and Qi, Haozhi and Malik, Jitendra and Pikul, James and Yim, Mark and Hellebrekers, Tess},
  booktitle={2025 IEEE International Conference on Robotics and Automation (ICRA)}, 
  title={Learning In-Hand Translation Using Tactile Skin with Shear and Normal Force Sensing}, 
  year={2025},
  volume={},
  number={},
  pages={5850-5856},
  doi={10.1109/ICRA55743.2025.11127974}
}

@ARTICLE{83-noseworthy2025forge,
  author={Noseworthy, Michael and Tang, Bingjie and Wen, Bowen and Handa, Ankur and Kessens, Chad and Roy, Nicholas and Fox, Dieter and Ramos, Fabio and Narang, Yashraj and Akinola, Iretiayo},
  journal={IEEE Robotics and Automation Letters}, 
  title={FORGE: Force-Guided Exploration for Robust Contact-Rich Manipulation Under Uncertainty}, 
  year={2025},
  volume={10},
  number={5},
  pages={4436-4443},
  doi={10.1109/LRA.2025.3551637}
}

@ARTICLE{aslam2025dartbot,
  author={Aslam, Shoaib and Kumar, Krish and Zhou, Pokuang and Yu, Hongyu and Wang, Michael Yu and She, Yu},
  journal={IEEE Transactions on Automation Science and Engineering}, 
  title={DartBot: Overhand Throwing of Deformable Objects With Tactile Sensing and Reinforcement Learning}, 
  year={2025},
  volume={22},
  number={},
  pages={13644-13661},
  doi={10.1109/TASE.2025.3556875}
}

@ARTICLE{zhao2024fots,
  author={Zhao, Yongqiang and Qian, Kun and Duan, Boyi and Luo, Shan},
  journal={IEEE Robotics and Automation Letters}, 
  title={FOTS: A Fast Optical Tactile Simulator for Sim2Real Learning of Tactile-Motor Robot Manipulation Skills}, 
  year={2024},
  volume={9},
  number={6},
  pages={5647-5654},
  doi={10.1109/LRA.2024.3396665}
}

@InProceedings{MimicTouch,
  title = 	 {MimicTouch: Leveraging Multi-modal Human Tactile Demonstrations for Contact-rich Manipulation},
  author =       {Yu, Kelin and Han, Yunhai and Wang, Qixian and Saxena, Vaibhav and Xu, Danfei and Zhao, Ye},
  booktitle = 	 {Proceedings of The 8th Conference on Robot Learning},
  pages = 	 {4844--4865},
  year = 	 {2025},
  editor = 	 {Agrawal, Pulkit and Kroemer, Oliver and Burgard, Wolfram},
  volume = 	 {270},
  series = 	 {Proceedings of Machine Learning Research},
  month = 	 {06--09 Nov},
  publisher =    {PMLR},
  url = 	 {https://proceedings.mlr.press/v270/yu25c.html},
}

@InProceedings{VT-Refine,
  title = 	 {VT-Refine: Learning Bimanual Assembly with Visuo-Tactile Feedback via Simulation Fine-Tuning},
  author =       {Huang, Binghao and Xu, Jie and Akinola, Iretiayo and Yang, Wei and Sundaralingam, Balakumar and O'Flaherty, Rowland and Fox, Dieter and Wang, Xiaolong and Mousavian, Arsalan and Chao, Yu-Wei and Li, Yunzhu},
  booktitle = 	 {Proceedings of The 9th Conference on Robot Learning},
  pages = 	 {2484--2500},
  year = 	 {2025},
  editor = 	 {Lim, Joseph and Song, Shuran and Park, Hae-Won},
  volume = 	 {305},
  series = 	 {Proceedings of Machine Learning Research},
  month = 	 {27--30 Sep},
  publisher =    {PMLR},
  url = 	 {https://proceedings.mlr.press/v305/huang25b.html},
}

@article{sharpa-apple,
  title={{Towards Human-Like Manipulation through RL-Augmented Teleoperation and Mixture-of-Dexterous-Experts VLA}},
  author={Tang, Tutian and Ji, Xingyu and Xing, Wanli and Hao, Ce and Xu, Wenqiang and Shao, Lin and Lu, Cewu and Yu, Qiaojun and Pang, Jiangmiao and Zhang, Kaifeng},
  journal={arXiv preprint arXiv:2603.08122},
  year={2026}
}

@inproceedings{zhao2025touch,
  title={Touch begins where vision ends: Generalizable policies for contact-rich manipulation},
  author={Zhao, Zifan and Haldar, Siddhant and Cui, Jinda and Pinto, Lerrel and Bhirangi, Raunaq},
  booktitle={3rd RSS Workshop on Dexterous Manipulation: Learning and Control with Diverse Data},
  year={2025}
}

@article{luu2025manifeel,
  title={{ManiFeel: Benchmarking and Understanding Visuotactile Manipulation Policy Learning}},
  author={Luu, Quan Khanh and Zhou, Pokuang and Xu, Zhengtong and Zhang, Zhiyuan and Qiu, Qiang and She, Yu},
  journal={arXiv preprint arXiv:2505.18472},
  year={2025}
}

@article{chen2026univtac,
  title={{UniVTAC: A Unified Simulation Platform for Visuo-Tactile Manipulation Data Generation, Learning, and Benchmarking}},
  author={Chen, Baijun and Wan, Weijie and Chen, Tianxing and Guo, Xianda and Xu, Congsheng and Qi, Yuanyang and Zhang, Haojie and Wu, Longyan and Xu, Tianling and Li, Zixuan and others},
  journal={arXiv preprint arXiv:2602.10093},
  year={2026}
}

@InProceedings{rodriguez2025cross,
  title = 	 {Cross-Sensor Touch Generation},
  author =       {Rodriguez, Samanta and Dou, Yiming and Oller, Miquel and Owens, Andrew and Fazeli, Nima},
  booktitle = 	 {Proceedings of The 9th Conference on Robot Learning},
  pages = 	 {152--167},
  year = 	 {2025},
  editor = 	 {Lim, Joseph and Song, Shuran and Park, Hae-Won},
  volume = 	 {305},
  series = 	 {Proceedings of Machine Learning Research},
  month = 	 {27--30 Sep},
  publisher =    {PMLR},
  url = 	 {https://proceedings.mlr.press/v305/rodriguez25a.html},
}

@article{chen2026training,
  title={Training tactile sensors to learn force sensing from each other},
  author={Chen, Zhuo and Ou, Ni and Zhang, Xuyang and Wu, Zhiyuan and Zhao, Yongqiang and Wang, Yupeng and Papastavridis, Emmanouil Spyrakos and Lepora, Nathan and Jamone, Lorenzo and Deng, Jiankang and others},
  journal={Nature Communications},
  year={2026},
  publisher={Nature Publishing Group UK London}
}

@InProceedings{pi0.5,
  title = 	 {$\pi_{0.5}$: a Vision-Language-Action Model with Open-World Generalization},
  author =       {Black, Kevin and Brown, Noah and Darpinian, James and Dhabalia, Karan and Driess, Danny and Esmail, Adnan and Equi, Michael Robert and Finn, Chelsea and Fusai, Niccolo and Galliker, Manuel Y. and Ghosh, Dibya and Groom, Lachy and Hausman, Karol and ichter, brian and Jakubczak, Szymon and Jones, Tim and Ke, Liyiming and LeBlanc, Devin and Levine, Sergey and Li-Bell, Adrian and Mothukuri, Mohith and Nair, Suraj and Pertsch, Karl and Ren, Allen Z. and Shi, Lucy Xiaoyang and Smith, Laura and Springenberg, Jost Tobias and Stachowicz, Kyle and Tanner, James and Vuong, Quan and Walke, Homer and Walling, Anna and Wang, Haohuan and Yu, Lili and Zhilinsky, Ury},
  booktitle = 	 {Proceedings of The 9th Conference on Robot Learning},
  pages = 	 {17--40},
  year = 	 {2025},
  editor = 	 {Lim, Joseph and Song, Shuran and Park, Hae-Won},
  volume = 	 {305},
  series = 	 {Proceedings of Machine Learning Research},
  month = 	 {27--30 Sep},
  publisher =    {PMLR},
  url = 	 {https://proceedings.mlr.press/v305/black25a.html},
}

@Inbook{villani2016force,
    author="Villani, Luigi
    and De Schutter, Joris",
    editor="Siciliano, Bruno
    and Khatib, Oussama",
    title="Force Control",
    bookTitle="Springer Handbook of Robotics",
    year="2016",
    publisher="Springer International Publishing",
    address="Cham",
    pages="195--220",
    isbn="978-3-319-32552-1",
    doi="10.1007/978-3-319-32552-1_9",
    url="https://doi.org/10.1007/978-3-319-32552-1_9"
}

@inproceedings{MoE,
  title={Outrageously Large Neural Networks: The Sparsely-Gated Mixture-of-Experts Layer},
  author={Shazeer, Noam and Mirhoseini, Azalia and Maziarz, Krzysztof and Davis, Andy and Le, Quoc and Hinton, Geoffrey and Dean, Jeff},
  booktitle={International Conference on Learning Representations},
  year={2017}
}

@article{lei2025rl,
    author = {Kun Lei  and Huanyu Li  and Dongjie Yu  and Zhenyu Wei  and Lingxiao Guo  and Zhennan Jiang  and Ziyu Wang  and Shiyu Liang  and Huazhe Xu },
    title = {Performant robotic manipulation with real-world reinforcement learning},
    journal = {Science Robotics},
    volume = {11},
    number = {116},
    pages = {eaed6267},
    year = {2026},
    doi = {10.1126/scirobotics.aed6267},
    URL = {https://www.science.org/doi/abs/10.1126/scirobotics.aed6267},
    eprint = {https://www.science.org/doi/pdf/10.1126/scirobotics.aed6267},
}

@article{chi2025diffusion,
    author = {Cheng Chi and Zhenjia Xu and Siyuan Feng and Eric Cousineau and Yilun Du and Benjamin Burchfiel and Russ Tedrake and Shuran Song},
    title ={Diffusion policy: Visuomotor policy learning via action diffusion},
    journal = {The International Journal of Robotics Research},
    volume = {44},
    number = {10-11},
    pages = {1684-1704},
    year = {2025},
    doi = {10.1177/02783649241273668},
    URL = {https://doi.org/10.1177/02783649241273668},
    eprint = {https://doi.org/10.1177/02783649241273668},
}

@INPROCEEDINGS{ACT, 
    AUTHOR    = {Tony Z. Zhao AND Vikash Kumar AND Sergey Levine AND Chelsea Finn}, 
    TITLE     = {{Learning Fine-Grained Bimanual Manipulation with Low-Cost Hardware}}, 
    BOOKTITLE = {Proceedings of Robotics: Science and Systems}, 
    YEAR      = {2023}, 
    ADDRESS   = {Daegu, Republic of Korea}, 
    MONTH     = {July}, 
    DOI       = {10.15607/RSS.2023.XIX.016} 
}

@INPROCEEDINGS{VINN, 
    AUTHOR    = {Jyothish Pari AND {Nur Muhammad (Mahi)} Shafiullah AND {Sridhar Pandian} Arunachalam AND Lerrel Pinto}, 
    TITLE     = {{The Surprising Effectiveness of Representation Learning for Visual Imitation}}, 
    BOOKTITLE = {Proceedings of Robotics: Science and Systems}, 
    YEAR      = {2022}, 
    ADDRESS   = {New York City, NY, USA}, 
    MONTH     = {June}, 
    DOI       = {10.15607/RSS.2022.XVIII.010} 
}

@InProceedings{RT-2,
  title = 	 {RT-2: Vision-Language-Action Models Transfer Web Knowledge to Robotic Control},
  author =       {Zitkovich, Brianna and Yu, Tianhe and Xu, Sichun and Xu, Peng and Xiao, Ted and Xia, Fei and Wu, Jialin and Wohlhart, Paul and Welker, Stefan and Wahid, Ayzaan and Vuong, Quan and Vanhoucke, Vincent and Tran, Huong and Soricut, Radu and Singh, Anikait and Singh, Jaspiar and Sermanet, Pierre and Sanketi, Pannag R. and Salazar, Grecia and Ryoo, Michael S. and Reymann, Krista and Rao, Kanishka and Pertsch, Karl and Mordatch, Igor and Michalewski, Henryk and Lu, Yao and Levine, Sergey and Lee, Lisa and Lee, Tsang-Wei Edward and Leal, Isabel and Kuang, Yuheng and Kalashnikov, Dmitry and Julian, Ryan and Joshi, Nikhil J. and Irpan, Alex and Ichter, Brian and Hsu, Jasmine and Herzog, Alexander and Hausman, Karol and Gopalakrishnan, Keerthana and Fu, Chuyuan and Florence, Pete and Finn, Chelsea and Dubey, Kumar Avinava and Driess, Danny and Ding, Tianli and Choromanski, Krzysztof Marcin and Chen, Xi and Chebotar, Yevgen and Carbajal, Justice and Brown, Noah and Brohan, Anthony and Arenas, Montserrat Gonzalez and Han, Kehang},
  booktitle = 	 {Proceedings of The 7th Conference on Robot Learning},
  pages = 	 {2165--2183},
  year = 	 {2023},
  editor = 	 {Tan, Jie and Toussaint, Marc and Darvish, Kourosh},
  volume = 	 {229},
  series = 	 {Proceedings of Machine Learning Research},
  month = 	 {06--09 Nov},
  publisher =    {PMLR},
  url = 	 {https://proceedings.mlr.press/v229/zitkovich23a.html},
}

@article{OpenAI-dexterous,
    author = {OpenAI: Marcin Andrychowicz and Bowen Baker and Maciek Chociej and Rafal Józefowicz and Bob McGrew and Jakub Pachocki and Arthur Petron and Matthias Plappert and Glenn Powell and Alex Ray and Jonas Schneider and Szymon Sidor and Josh Tobin and Peter Welinder and Lilian Weng and Wojciech Zaremba},
    title ={Learning dexterous in-hand manipulation},
    journal = {The International Journal of Robotics Research},
    volume = {39},
    number = {1},
    pages = {3-20},
    year = {2020},
    doi = {10.1177/0278364919887447},
    URL = {https://doi.org/10.1177/0278364919887447},
    eprint = {https://doi.org/10.1177/0278364919887447},
}

@article{johansson2009coding,
  title={Coding and use of tactile signals from the fingertips in object manipulation tasks},
  author={Johansson, Roland S and Flanagan, J Randall},
  journal={Nature Reviews Neuroscience},
  volume={10},
  number={5},
  pages={345--359},
  year={2009},
  publisher={Nature Publishing Group UK London}
}

\end{document}